%% file: main_20240503174405.tex
\documentclass[runningheads]{llncs}
\usepackage[T1]{fontenc}

\PassOptionsToPackage{numbers}{natbib}
\input{math_commands.tex}

\usepackage[utf8]{inputenc} %
\usepackage[hidelinks]{hyperref}       %
\usepackage{url}            %

\usepackage{wrapfig}
\usepackage{authblk}
\usepackage[symbol]{footmisc}
\usepackage{graphicx}
\usepackage{multirow}
\usepackage{multicol}   
\usepackage{color, colortbl, xcolor}
\usepackage{caption}

\usepackage{graphicx}

\def\arrvline{\hfil\kern\arraycolsep\vline\kern-\arraycolsep\hfilneg}
\usepackage{booktabs}       %
\usepackage{amsfonts}       %
\usepackage{nicefrac}       %
\usepackage{microtype}      %
\usepackage{xcolor}         %
\usepackage{lmodern}

\begin{document}

\title{A Fresh Look at Sanity Checks for Saliency Maps}

\titlerunning{A Fresh Look at Sanity Checks for Saliency Maps}
\author{Anna Hedström\inst{1,2,3,*}\orcidID{0009-0007-7431-7923} \and
Leander Weber\inst{2,*}\orcidID{0000-0003-2995-6914} \and
Sebastian Lapuschkin\inst{2}\orcidID{0000-0002-0762-7258} \and 
Marina Höhne\inst{3,4,5}\orcidID{0000-0003-3090-6279}}
\authorrunning{A. Hedström et al.}
\institute{Department of Electrical Engineering and Computer Science, TU Berlin \and
Department of Artificial Intelligence, Fraunhofer HHI, Berlin, Germany \and
UMI Lab, Leibniz Institute of Agricultural Engineering and Bioeconomy e.V. (ATB) \and
BIFOLD – Berlin Institute for the Foundations of Learning and Data \and 
Department of Computer Science,
University of Potsdam \\ \email{sebastian.lapuschkin@hhi.fraunhofer.de}, \email{MHoehne@atb-potsdam.de}}
\maketitle            
\def\thefootnote{*}\footnotetext{Equal Contribution}\def\thefootnote{\arabic{footnote}}

\def\thefootnote{\arabic{footnote}}
\setcounter{footnote}{0}

\begin{abstract}
The Model Parameter Randomisation Test (MPRT) is highly recognised in the eXplainable Artificial Intelligence (XAI) community due to its fundamental evaluative criterion: explanations should be sensitive to the parameters of the model they seek to explain. 
However, recent studies have raised several methodological concerns for the empirical interpretation of MPRT.
In response, we propose two modifications to the original test: \textit{Smooth MPRT} and \textit{Efficient MPRT}. The former reduces the impact of noise on evaluation outcomes via sampling, while the latter avoids the need for biased similarity measurements by re-interpreting the test through the increase in explanation complexity after full model randomisation. Our experiments show that these modifications enhance the metric reliability, facilitating a more trustworthy deployment of explanation methods.

\keywords{Explainability \and Evaluation \and Faithfulness \and Quantification}
\end{abstract}
\section{Introduction}

Evaluating the quality of explanation methods remains a challenge in the field of eXplainable Artificial Intelligence (XAI), because generally, ground truth explanation labels do not exist~\cite{Bellido1993,dasgupta2022,hedstrom2022quantus}. To address this issue, various quantitative evaluation methods have been proposed~\cite{montavon2018,alvarezmelis2018robust,yeh2019,dasgupta2022,bhatt2020,alvarezmelis2018robust,agarwalstability,nguyen2020,bach2015pixel,samek2015,ancona2019,irof2020,yeh2019,rong2022,dasgupta2022,adebayo2018}, among these the \textit{Model Parameter Randomisation Test} (MPRT)~\cite{adebayo2018}.
This test quantifies how faithfully an explanation method explains the model by randomising the model's parameters, starting with the output layer and progressing backward. According to this test, a more significant change in explanations during the progressive randomisation of the model indicates higher explanation quality.

\begin{figure}[!t]
    \centering
    \includegraphics[width=\linewidth]{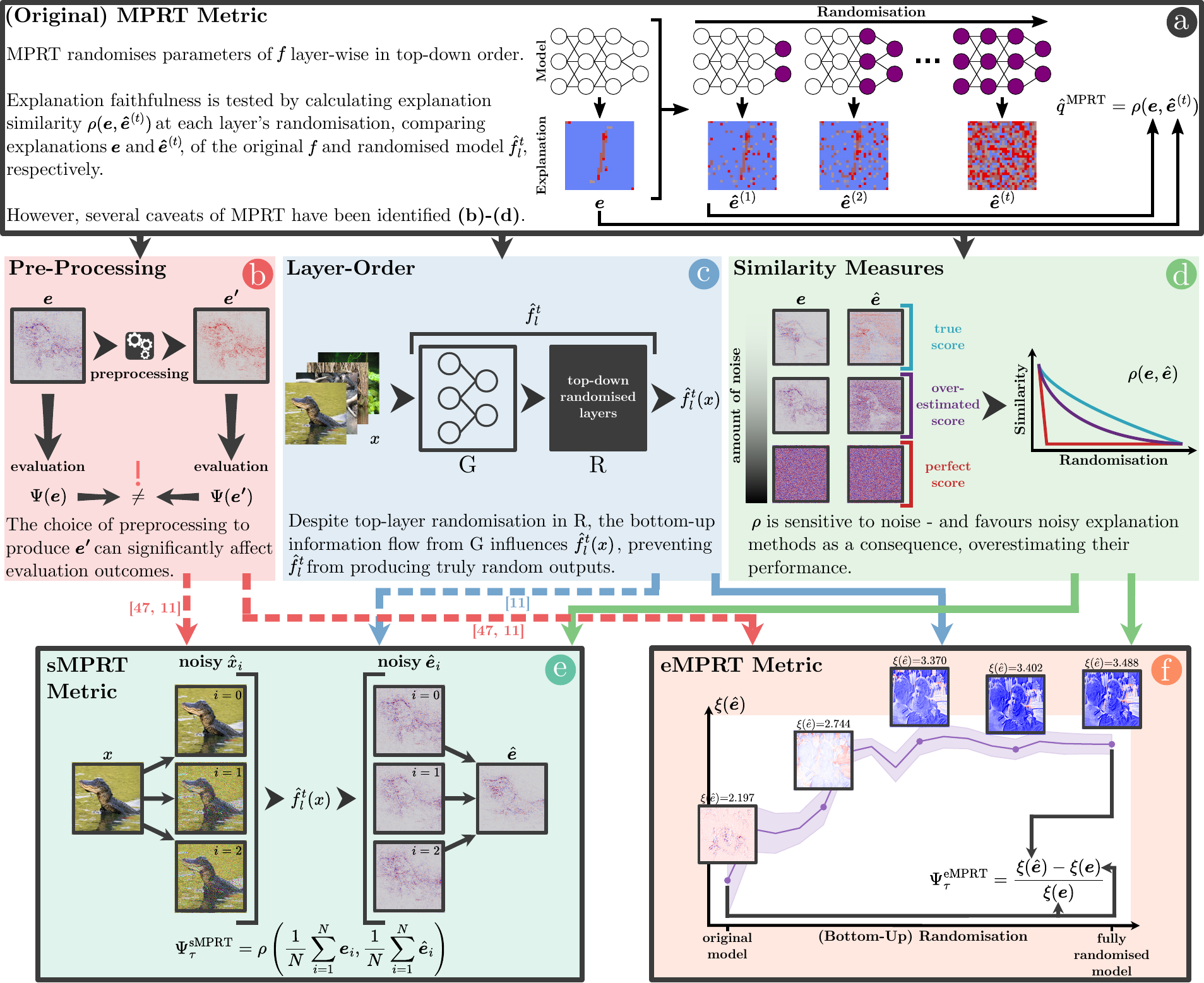}
    \captionsetup{font=footnotesize}
    \caption{
    Schematic visualisation of the original MPRT~\cite{adebayo2018} (\emph{top}), shortcomings identified by recent literature (\emph{middle}) and our proposed solutions (\emph{bottom}). Solid arrows in the visualization indicate shortcomings directly addressed by our proposed metrics. Dashed arrows show those resolved through incorporating methods suggested by existing research~\cite{sundararajan2018,bindershort}. (a) The MPRT assesses the reliability of explanation methods by randomising a model's parameters layer by layer and comparing explanation similarity, i.e., $\rho(\ve, \hat{\ve})$ between the original model $f$ and the randomised version $\hat{f}$. This is done by examining the explanations from each layer. (b) \textit{Pre-processing}: normalisation and using absolute values can affect MPRT results by stripping information related to feature importance, especially the sign.
    (c) \textit{Layer-order}: randomising layers from top to bottom can retain properties of the original lower layers and not yield fully random outputs, skewing the evaluation. (d) \textit{Similarity measures}: the pairwise similarity metrics used in the original MPRT are sensitive to noise (e.g. from gradient shattering), potentially affecting how the test ranks explanation methods. (e) sMPRT introduces a ``denoising'' pre-processing step that averages explanations over $N$ perturbed inputs, reducing noise.
    (f) eMPRT reinterprets MPRT by comparing the complexity, measured by discrete entropy $\xi(\ve)$, of explanations before and after full model randomisation. }
    \label{fig:overview}
    \vspace{-0.75em}
\end{figure}

In recent years, several independent works have raised questions about methodological aspects of the original MPRT~\cite{sundararajan2018,kokhlikyan2021investigating,yona2021,bindershort}. These concerns point out various aspects of the test, such as the pairwise similarity measure used to evaluate changes in explanations~\cite{bindershort}, the order in which model layers are randomised~\cite{bindershort}, the preprocessing of explanation outputs~\cite{sundararajan2018,bindershort}, and the influence of the specific model and task~\cite{yona2021,kokhlikyan2021investigating}. Given the widespread use of MPRT within the XAI community and its potential impact on the rejection or acceptance of different explanation methods \cite{sixt2019,GHASSEMI2021e745,Luxburg24}, these issues raise an important question: does the original MPRT effectively evaluate the quality of an explanation method, or are there ways to enhance its reliability? 

The fundamental concept underpinning MPRT, which asserts that the explanation function ought to be responsive to alterations in the model's parameters, stands as both important and insightful. However, known shortcomings, such as the order of randomising model layers and the choice of pairwise similarity measures, as outlined in Section \ref{issues} and shown in Figure \ref{fig:overview}, should be addressed. Building upon this initial work, we therefore propose two new variants of the MPRT: \textit{Smooth MPRT} (sMPRT) and \textit{Efficient MPRT} (eMPRT). The sMPRT reduces sensitivity to shattering noise (cf.~\cite{bindershort}) through denoising of explanations, achieved by averaging the attribution results across $N$ perturbed inputs. The eMPRT measures the faithfulness of explanations by quantifying their increase in complexity after full parameter randomisation, thereby eliminating reliance on potentially biased similarity measures. Both extensions of MPRT are made publicly available as part of the Quantus\footnote{Code at: \url{https://github.com/understandable-machine-intelligence-lab/Quantus}.} evaluation library~\cite{hedstrom2022quantus}. 

While our proposed methods are iterative in nature, they carry substantial importance. When deploying XAI in real-world applications, particularly in safety-critical areas, the emphasis must be on creating robust quantitative evaluation metrics to ensure that explanation methods are of sufficient quality (or at least, pass sanity checks). Our work contributes to ensuring that the development of XAI methods is not only theoretically grounded but also empirically valid. 

\section{Preliminaries}

We first introduce the core notation and subsequently define MPRT. %

\textbf{Local Explanations.} Let $f : \mathcal{X} \mapsto \mathcal{Y}$, $f \in \mathcal{F}$ be a black-box, neural network model that maps inputs $\vx \in \mathbb{R}^D$ to predictions $\vy \in \mathbb{R}^C$ with $C$ classes, where $\mathcal{F}$ denotes the function space. We assume that $f$ consists of $L$ layers $f^1, \cdots, f^L$, such that $f(\vx) = f^{L} \circ \dots \circ f^{1}(\vx)$.  To interpret the behaviour of $f$, \emph{local} explanation methods~\cite{smilkov2017smoothgrad,ZeilerOcclusion,sundararajan2017axiomatic,selvaraju2019,bykov2021noisegrad} attribute importance scores to features of input $\vx$, given a prediction of a class $c \in [1, C]$ such that $y := y_c$. We define an explanation ${\ve} \in \mathbb{R}^D$ via the explanation function $\Phi_{\lambda}: \mathcal{F} 
 \times \mathcal{X} \times \mathcal{Y} \mapsto \mathbb{R}^D$ with $\ve = \Phi(f, \vx, y; \lambda)$. Then $\mathcal{E}$ denotes the space of possible explanations w.r.t. $\lambda$ with $\Phi_{\lambda} \in \mathbb{E}$. Here, $\lambda$ denotes explanation method-specific parameters. 

\textbf{Evaluating Explanations.} Evaluating the quality of explanations $\Phi$ for black-box models is challenging~\cite{Bellido1993,bbox1997,dasgupta2022,hedstrom2023metaquantus} due to the absence of ground truth. Instead, quality is usually measured through specific properties of explanations, such as stability (e.g.~\cite{montavon2018,alvarezmelis2018robust,yeh2019}), complexity (e.g.,~\cite{nguyen2020,bhatt2020,chalasani2020}) or faithfulness (e.g.,~\cite{adebayo2018,bhatt2020,samek2015}). For an overview of existing evaluation properties, see previous works~\cite{hedstrom2022quantus,hedstrom2023metaquantus}.
We define a quality estimate ${q} \in \mathbb{R}$ as the output of an evaluation function $\Psi_{\tau}: \mathcal{E} \times \mathcal{F} 
 \times \mathcal{X} \times \mathcal{Y} \mapsto \mathbb{R}$, parameterised by $\tau$ such that $q = \Psi_{\tau}(\Phi,f, \vx, y)$. %
Then $\sO$ denotes the space of possible evaluations w.r.t. $\ve$ and $\Psi \in \sO$. Here, $\tau$ denotes explanation method-specific parameters.

\subsection{Model Parameter Randomisation Test (MPRT)}
\label{sec:mprt}

We define the original MPRT~\cite{adebayo2018} as follows:
\begin{theorem}[MPRT]
\label{def-mprt}
\textit{Let $\Psi_{\tau}^{\text{MPRT}}: \mathcal{E} \times \mathcal{F} 
 \times \mathcal{X} \times \mathcal{Y} \mapsto \mathbb{R}$ be an evaluation function that computes a quality estimate $\hat{q} \in \mathbb{R}$ that measures the similarity between the original explanation $\ve_l$ and the explanation $\hat{\ve} := \Phi(\vx, \hat{f}^{t}_l, y)$ corresponding to the perturbed model $\hat{f}^{t}_l$ randomised in a top-down fashion up to layer $l \in [L, L-1, \ldots, 1]$:} 
\begin{equation}\label{eq:q_mprt}
\hat{q}^{\text{MPRT}} = \rho(\ve, \hat{\ve}_l),
\end{equation}
\textit{with $\rho: \mathbb{R}^{D} \times \mathbb{R}^{D} \mapsto \mathbb{R}$ as a similarity function.}
\end{theorem}
 
The original MPRT thus progressively randomises model layers top-down (from $L$ to $1$). This is denoted by $\hat{f}_l^{t}$, where $l \in [L, L-1, \ldots, 1]$ represents the layer index. Alternatively, for bottom-up randomisation, we denote $\hat{f}_l^{b}$ with $l \in [1, 2, \ldots, L]$. The test compares explanations of the randomised and unrandomised model at each step to evaluate explanation faithfulness. The seminal work~\cite{adebayo2018} posits that the quicker similarity $\hat{q}^{\text{MPRT}}$ decreases with randomisation, the more faithful the explanation method is. This means that if the explanation function $\Phi$ is sensitive to model parameters, the original- $\ve$ and the randomised explanation $\hat{\ve}_l$ for a given class $y$ should not be similar, i.e., $\hat{q}^{\text{MPRT}} \ll 1$. 

\subsection{Methodological Caveats} \label{issues} %

Several methodological concerns have been raised about the MPRT, particularly regarding its use of (a) pre-processing, (b) layer-order, and (c) similarity measures. In the following, we summarise these caveats and proposed solutions:

\textbf{(a) Pre-processing.} In the initial implementation of MPRT~\cite{adebayo2018}, the normalisation of explanations $\ve$ and $\hat{\ve}$ was done using their minimum and maximum values. This approach is problematic because these these statistics are highly variable and almost arbitrary across different attribution methods, which confounds comparative analyses across evaluations~\cite{bindershort}. The use of absolute values for attributions can remove valuable information~\cite{sundararajan2018} --- a limitation already highlighted in the original work~\cite{adebayo2018}. \textit{(Proposed Solution) To preserve the scale and distribution of explanations, the authors of~\cite{bindershort} suggest to normalise $\ve$ using the square root of the \textit{average second-moment estimate} (see Equation \ref{eq:normalise-eq} in Supplements). This method avoids constraining attributions into a fixed range, facilitating comparability across various randomisations and methods~\cite{bindershort}.}

\textbf{(b) Layer-order.} While it might seem logical to expect a significant change in the explanation when randomising layers top-down, i.e., $\rho(\ve, \hat{\ve}_l) \ll 1$ over layers $l \in [L, L-1, \ldots, 1]$, recent research suggests otherwise~\cite{bindershort}. As discussed by the authors of~\cite{bindershort}, randomising the model's top layers produces only minor changes in the forward pass due to several reasons: (i) irrelevant features from non-randomised lower layers persist into higher, randomised layers; (ii) high activations in the earlier layers may continue to dominate the network's response, even after randomisation; and (iii) architectures with skip connections, such as ResNets~\cite{he2015deep}, retain certain constant baseline explanations due to their structure that partially preserves features in the forward pass. 
Thus, even after randomising the top layers, the expected degree of change between $\ve$ and $\hat{\ve}_l$ is limited, suggesting that faithful explanation methods may not show significant alterations after model randomisation.
\textit{(Proposed Solution) To prevent the retention of information during the forward pass, it is preferable to randomise layers from the bottom-up (as discussed in Supplement \ref{sec:layerorder_supplement}), or to assess the change in explanations only after complete randomisation of all layers (see Equation \ref{eq:q_emprt}.}

\textbf{(c) Similarity Measures.} Another limitation identified by the authors of~\cite{bindershort} relates to the use of pairwise similarity measures such as the \textit{Structural Similarity Index} (SSIM)~\cite{nilsson2020understanding} and \textit{Spearman Rank Correlation}~\cite{spearman1961proof}, which can be minimised by statistically uncorrelated random processes. This issue raises concerns because it suggests that some explanation methods, particularly those with inherent (shattering) noise, e.g., gradient-based techniques, appear more favourable in evaluations, leading to biased results in comparisons and rankings of explanation methods. \textit{(Proposed Solution) To address these shortcomings, we propose two modifications: sMPRT and eMPRT. These approaches address the limitations from different angles where sMPRT mitigates the impact of noise by applying a denoising step for pre-processing explanations and eMPRT replaces pairwise similarity measures with a complexity measure.}

\section{Methods}

In the following, we provide mathematical details and motivations for sMPRT and eMPRT.

\subsection{Smooth Model Parameter Randomisation Test (sMPRT)}
\label{sec:smprt}

The first method we propose is the sMPRT, which mitigates shattering noise found in some local explanation methods~\cite{tessahan2022} via a small modification of the original MPRT. Drawing inspiration from earlier approaches~\cite{smilkov2017smoothgrad,bykov2021noisegrad}, sMPRT introduces a preprocessing step to the evaluation process. Given an input $\vx$, it generates $N$ perturbed instances $\hat{\vx}_i$, computes the corresponding attributions $\Phi(\hat{\vx}_i, f, y; \lambda)$ with $i \in [1, N]$, and then applies the MPRT evaluation to the averaged, denoised attribution estimates.
We define the sMPRT in the following.
\begin{theorem}[sMPRT]
\label{def-smprt}
\textit{Let $\Psi_{\tau}^{\text{sMPRT}}: \mathcal{E} \times \mathcal{F} 
 \times \mathcal{X} \times \mathcal{Y} \mapsto \mathbb{R}$ be an evaluation function that computes a quality estimate $\hat{q} \in \mathbb{R}$ that measures the similarity estimates between explanations $\ve_i := \Phi(\hat{\vx}_i, f, y; \lambda)$ and $\hat{\ve}_{l, i} := \Phi(\hat{\vx}_i, \hat{f}_l^b, y; \lambda)$ averaged over $i \in [1, N]$ where $\hat{\ve}_{l, i}$ corresponds to the perturbed model $\hat{f}^{b}_l$ randomised in a bottom-down fashion up to layer $l \in [L, L-1, \ldots, 1]$:}
\begin{equation}\label{eq:q_smprt}
\hat{q}^{\text{sMPRT}} = \rho\left(\frac{1}{N} \sum_{i=1}^N \ve_i, \frac{1}{N} \sum_{i=1}^N \hat{\ve}_{l, i}\right),
\end{equation}
\textit{where $\hat{\vx}_i = \vx + \eta_i$, $\eta_i \sim \mathcal{N}(0, \sigma)$ and $||\eta_i||_p \leq \epsilon$ holds with high probability, with $\sigma, \epsilon \in \mathbb{R}$.}
\end{theorem}
Similar to Equation \ref{eq:q_mprt}, values such that $\hat{q}^{\text{sMPRT}} \ll 1$ is expected from Equation \ref{eq:q_smprt}, implying low similarity between $\ve$ and $\hat{\ve}_l$. To set $\sigma$, we follow the heuristic provided in the original publication~\cite{smilkov2017smoothgrad}, i.e., $\sigma/(x_\text{max} - x_\text{min})$, as detailed in the Supplement. We investigate multiple values for $N$ in Section \ref{sec:smprt_neffect}, but find that $N=50$ seems to best balance consistent results and computational expense.

\subsection{Efficient Model Parameter Randomisation Test (eMPRT)}
\label{sec:emprt}
To overcome limitations (b) and (c) (Section \ref{issues}), while also enhancing the efficiency of MPRT, we propose eMPRT. This test eliminates the need for layer-by-layer comparisons between $\ve$ and $\hat{\ve}_l$, and instead examines the \textit{relative rise in explanation complexity} between the original- and the fully randomised model.
We define eMPRT in the following.
\begin{theorem}[eMPRT]
\label{def-emprt}
\textit{Let $\Psi_{\tau}^{\text{eMPRT}}: \mathcal{E} \times \mathcal{F} 
 \times \mathcal{X} \times \mathcal{Y} \mapsto \mathbb{R}$ be an evaluation function that computes a quality estimate $\hat{q} \in \mathbb{R}$ that measures the relative rise in the complexity of the explanation from a fully randomised model $\hat{f}$ such that $\hat{\ve} := \Phi(\vx, \hat{f}, y; \lambda)$:}
\begin{equation}\label{eq:q_emprt}
\hat{q}^{\text{eMPRT}} = \frac{\xi(\hat{\ve}) - \xi(\ve)}{\xi(\ve)}
\end{equation}
\textit{where $\xi: \mathbb{R}^{D} \mapsto \mathbb{R}$ is a complexity function, e.g., defined in Equation \ref{eq:hist_entropy}.}
\end{theorem}

For $\Phi$ to be faithful w.r.t $f$, we would expect that a fully randomised model, with higher entropy at $f_L(\vx)$ (as shown in Figure \ref{fig:emprt}(e)), yields a explanation $\hat{\ve}$ of reduced information value. Equation \ref{eq:q_emprt} capture this expectation as a rise in relative entropy. Positive values, i.e., $\hat{q}^{\text{eMPRT}} > 0$ then imply a rise in complexity, while negative values, i.e., $\hat{q}^{\text{eMPRT}} < 0$ suggests a decreasing complexity. $\hat{q}^{\text{eMPRT}} = 0$ denotes a neutral value, with unchanging explanation complexity.

We measure $\xi$ using the entropy of a histogram, grounded in Shannon-Entropy~\cite{Shannon48}. Here, we first bin attribution values $\ve$ over $B$ distinct bins where the frequency $c_i$ is measured for each $i^{\text{th}}$ bin, and then compute the corresponding normalised probability, $p_i$ as follows:
\begin{equation} \label{eq:hist_entropy}
\xi(\ve) = -\sum_{i=1}^B p_i \log(p_i) \quad \text{where} \quad p_i = \frac{c_i}{\sum_{i=1}^B c_i}.
\end{equation}
Using Equation \ref{eq:hist_entropy} in place of traditional entropy calculations, offers three key advantages. It retains the original sign of the attributions, provides implicit normalisation and is adaptable to different dimensionalities and distributions, wherein the parameter $B$ can be contextually chosen.

\section{Experimental Results}

The following section presents our results. Experimental details and additional experiments are found in Supplementary Material (Sections \ref{sec:experimental_setup_supplement}-\ref{sec:benchmarking-extra-results}). To replicate the experiments, please see code available at the GitHub repository\footnote{Experiments at: \url{https://github.com/annahedstroem/sanity-checks-revisited}.}.

\subsection{Analysing sMPRT}

In the following, we analyse the performance of sMPRT from two aspects: how effectively it denoises attributions during evaluation, and how many samples are necessary for a reliable quality estimate.

\begin{figure}[!b]
\centering    \includegraphics[width=\linewidth]{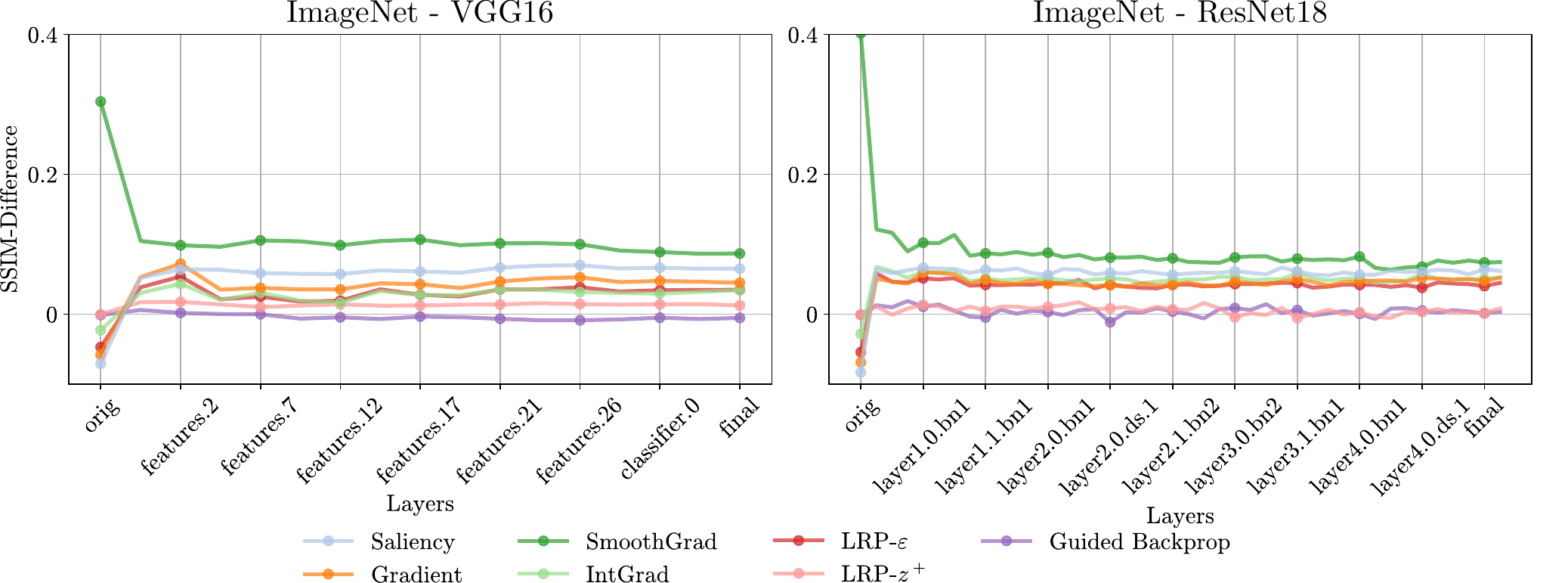}
    \captionsetup{font=footnotesize}
    \caption{Difference between sMPRT ($N=50$) and MPRT (corresponding to sMPRT with $N=1$). \textit{SSIM} performance can degrade when attributions are denoised by sMPRT, and this effect occurs most strongly with gradient-based methods such as \emph{Gradient} or \emph{SmoothGrad}. The unrandomised and fully randomised model states are denoted as \emph{orig} and \emph{final}, respectively.}
    \label{fig:smprt}
    \vspace{-0.5em}
\end{figure}

\subsubsection{Denoising Attributions with sMPRT.} 

Figure \ref{fig:smprt} visualises the difference in SSIM scores between sMPRT ($N=50$) and MPRT (which corresponds to sMPRT with $N=1$) over the course of bottom-up randomisation. Figure \ref{fig:smprt} visualises results for two different models, VGG-16~\cite{simonyan2014very} and ResNet-18~\cite{he2015deep} models, the ImageNet~\cite{ILSVRC15} dataset, and seven attribution methods. Please find Figure \ref{fig:smprt-data} for the raw sMPRT curves. 

Generally, increasing $N$ to 50 with sMPRT clearly impacts evaluation, although the amount of change varies between attribution methods. Specifically, gradient-based attribution methods (e.g., SmoothGrad~\cite{smilkov2017smoothgrad} and Saliency~\cite{Simonyandeep}) perform worse under sMPRT, with their SSIM scores strongly increasing. Modified backpropagation methods (e.g. \mbox{LRP-$z^+$}~\cite{bach2015pixel,montavon2017dtd}, Guided Backpropagation~\cite{springenberg2014striving}) are barely affected. This finding implies that after ``removing noise by adding noise'' (cf. Figure \ref{fig:smprt-data}), the performance discrepancy between explanation methods found in~\cite{adebayo2018} is reduced. Specifically, the maximum difference in SSIM between attribution methods over randomisation decreases from $\sim0.38$ (MPRT) to $\sim0.34$ (sMPRT) for VGG-16 and from $\sim0.50$ (MPRT) to $\sim0.46$ (sMPRT) for ResNet-18. This observation complements the theoretical findings of recent work~\cite{bindershort}, which argues that the similarity measures utilised by MPRT may be skewed and favour (noisy) gradient-based explanations. 

\subsubsection{Effect of the Number of Perturbed Samples.} 
\label{sec:smprt_neffect}

In this section, we examine how varying the number of perturbed samples, i.e., $N$ influences quality estimation with sMPRT. For comparability across different values of $N$, we compute the area under the curve (AUC) for each sMPRT curve. A lower AUC implies greater explanation faithfulness to the parameters of the model.

\begin{figure}[!t]
    \centering
\includegraphics[width=0.95\linewidth]{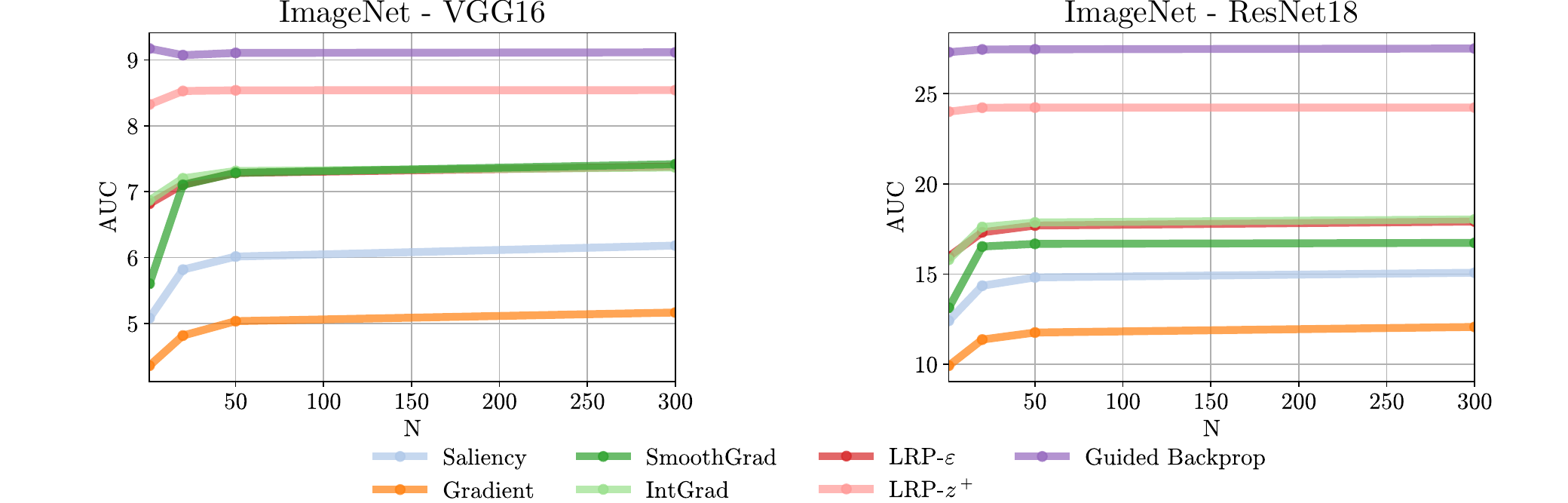}
    \captionsetup{font=footnotesize}
    \caption{Effect of number of perturbed samples $N$ on sMPRT results for VGG16 \emph{(left)} and ResNet18 \emph{(right)} on ImageNet data. The plots indicate how the area under the mean sMPRT curve (AUC) changes with $N$ for different explanation methods (i.e. the area under the curves as shown in Figure \ref{fig:smprt}). Up to $N=50$, there seems to be a steep change in AUC, especially for gradient-based methods. After that, the AUC curves flatten out, indicating a converged estimate of the denoised samples.}
    \label{fig:smprt_n}
    \vspace{-0.5em}
\end{figure}

Figure \ref{fig:smprt_n} illustrates how these AUC values change when $N$ increases. Gradient-based explanation methods tend to show more variation with changes in $N$, but the AUC tends to converge as $N$ increases, suggesting that the denoised sample estimate also stabilises with larger $N$. While even with $N=300$, slight variations occur, the AUC doesn't change substantially after reaching $N=50$. Consequently, considering the computational trade-off incurred by larger $N$, we recommend using $N=50$ when estimating explanation quality. While a larger $N$ will provide a more accurate estimate, it comes at increased computational cost.

\subsection{Analysing eMPRT}

In the following, we first discuss the eMPRT benchmarking results. Then, we investigate how categorical ranking outcomes differ, compared to MPRT.

\subsubsection{Benchmarking Explanation Methods.} 
\begin{figure}[!t]
    \centering
    \includegraphics[width=\linewidth]{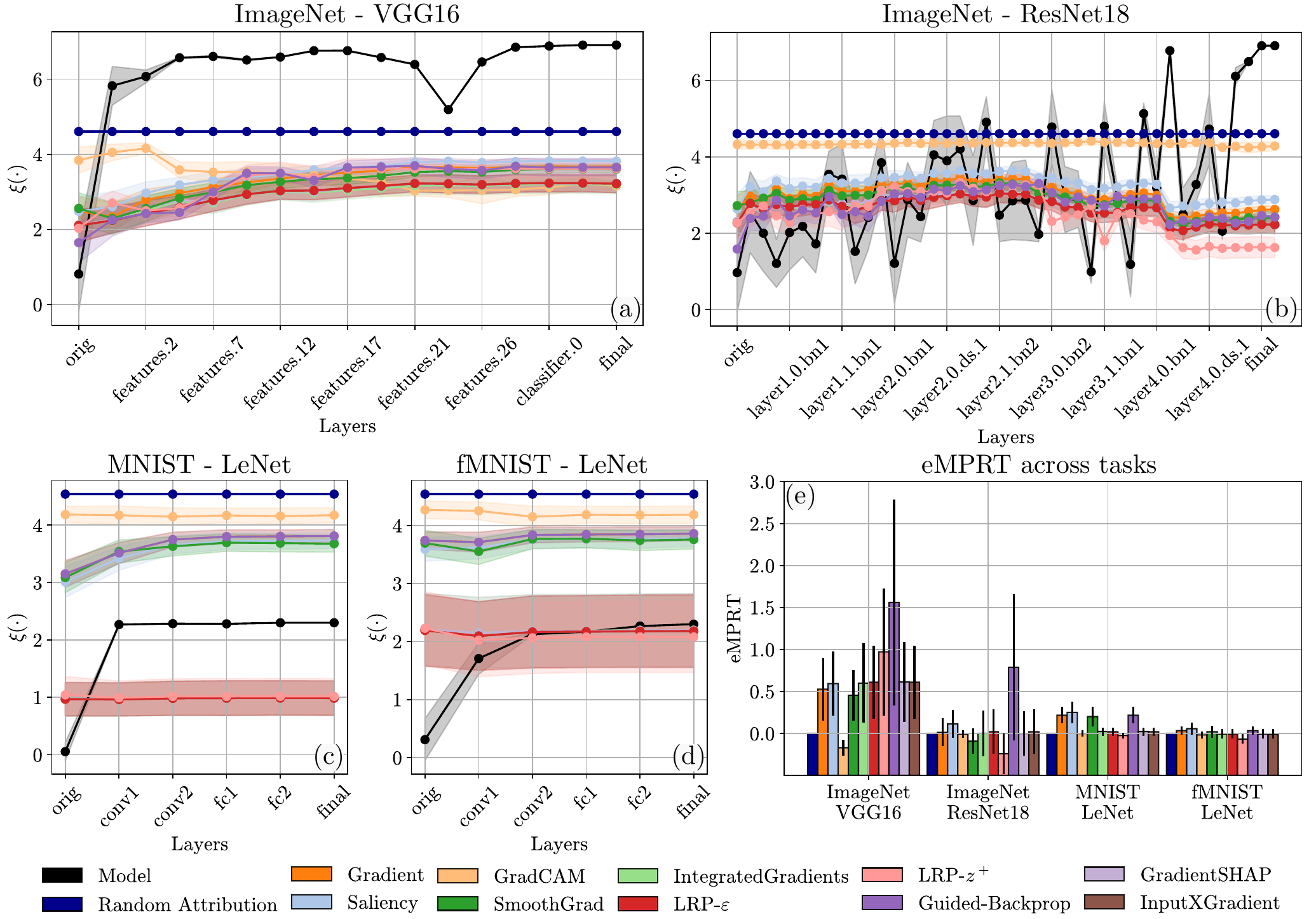}
    \captionsetup{font=footnotesize}
    \caption{Panels (a)-(d) illustrate entropy curves, representing the increase in complexity $\xi(\hat{e})$ with progressive bottom-up layer randomisation, denoted as $\hat{f}_l^b$. Panel (e) presents aggregated eMPRT scores, serving as a comparative benchmark for ten different XAI methods. No definitive superiority is observed across the tested datasets and models.}%
    \label{fig:emprt}
    \vspace{-1em}
\end{figure}

Figure \ref{fig:emprt} displays curves of discrete entropy, denoted as $\xi$, in panels (a) through (d), showing progressive randomisation of layers, and consolidated eMPRT scores in panel (e), spanning various tasks including VGG-16 and ResNet-18 models for ImageNet dataset, alongside LeNet for both MNIST~\cite{lecun2010mnist} and fMNIST~\cite{fashionmnist2015} datasets. This analysis includes ten different attribution methods and establishes a random attribution as a comparative baseline.  

Figure \ref{fig:emprt}(a)-(d) displays that even with full model randomisation, none of the methods reaches the theoretical randomness limit \emph{(dark blue)} that would indicate perfect faithfulness to ${f}$.
The progression of eMPRT curves exhibits considerable variation across different tasks. Such variation, anticipated due to architectural characteristics like the skip connections in ResNets, can complicate a practitioner's ability to assess the faithfulness of explanation methods through different layers. To ease the functional interpretation of quality estimation results, we propose to evaluate explanations solely based on their performance before and after full model randomisation (see Equation \ref{eq:q_emprt}).

By assessing the model's complexity (illustrated in \emph{black}, calculated from the non-discretised entropy of the model's output layer post-softmax, as shown in Equation \ref{shannon-entropy}) alongside the complexities of various explanations in Figure \ref{fig:emprt}(a)-(d), we can anchor our expectation of how the complexity of an ideal explanation, i.e., ``true to its model''~\cite{sturmfels2019}, ought to develop. Notably, none of the explanation methods effectively replicate the complexity of the model's function. This outcome is arguably unsurprising given the ``complexity gap'' between the functions involved, i.e., $f$ and $\Phi$, where the model parameters' dimensionality far surpasses the explanation functions' dimensionality or algorithmic complexity \cite{Luxburg24}.
Furthermore, Figure \ref{fig:emprt}(e) shows that no single method consistently surpasses the others across tested tasks.
Based on these observations, it is reasonable to conclude that all evaluated explanation methods underperform in absolute terms.

\begin{figure}[!t]
    \centering
    \includegraphics[width=0.75\linewidth]{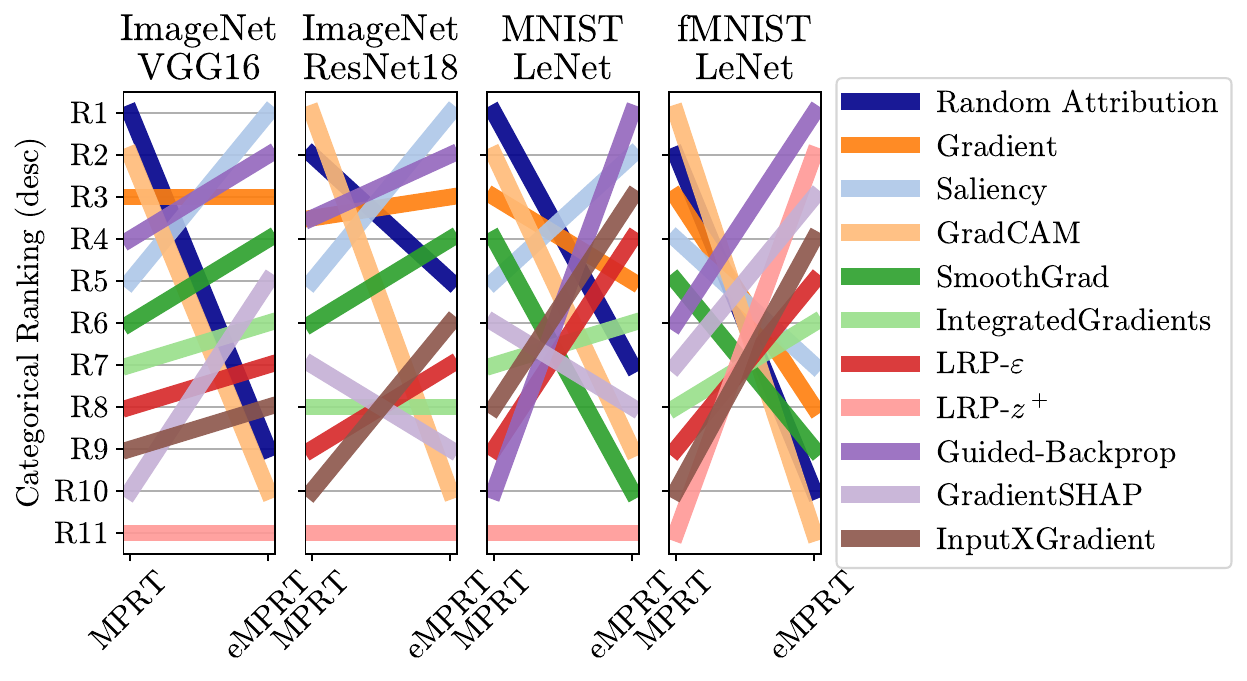}
    \captionsetup{font=footnotesize}
    \caption{Relative ranking of attribution methods using MPRT and eMPRT across different datasets and models. This figure illustrates the categorical rankings for ten attribution methods plus a random attribution across ImageNet (VGG16, ResNet18) and MNIST (LeNet) datasets. The evaluation reveals how rankings fluctuate strongly between MPRT and eMPRT, consistent with findings from~\cite{hedstrom2023metaquantus}.}
    \label{fig:emprt-vs-mprt-ranking}
\end{figure}

\subsubsection{Comparing Evaluation Outcomes with MPRT.} \label{sec:compare-rankings}

To understand the practical implications of using eMPRT versus MPRT, we evaluate the extent to which the evaluation rankings of explanation methods differ between these metric variants.

In Figure \ref{fig:emprt-vs-mprt-ranking}, we display the categorical rankings for ten different attribution methods, including a random attribution, as assessed by the MPRT and eMPRT methodologies. The rankings are presented in descending order, with $R1$ representing the highest performance and $R11$ the lowest. These results demonstrate considerable variability in explanation rankings between the variants. Contrary to assertions~\cite{adebayo2018} that claimed {Guided Backpropagation}~\cite{springenberg2014striving} to perform worse than gradient-based methods such as {Gradient}~\cite{morch,baehrens} and {SmoothGrad}~\cite{smilkov2017smoothgrad}, our eMPRT evaluations advance {Guided Backpropagation}~\cite{springenberg2014striving} above its gradient-based counterparts.

Figure \ref{fig:emprt-vs-mprt-ranking} further highlights how both metrics score random attributions, which ideally should receive low scores to reflect their lack of meaning~\cite{dasgupta2022}. Interestingly, the random attribution method, which serves as a lower bound for explanation faithfulness, consistently scores lower under eMPRT than MPRT, indicating an advantage with eMPRT evaluations. It is also worth noting that random attribution does not always receive the lowest rankings; in some settings, LRP-$\varepsilon$~\cite{bach2015pixel,montavon2017dtd}, GradCAM~\cite{selvaraju2019}, and SmoothGrad~\cite{smilkov2017smoothgrad} are ranked lower, albeit with small margins.

\subsection{Meta-Evaluation} 

It is critical to explore how the variants MPRT, sMPRT, and eMPRT differ in their metric performance characteristics. To achieve this, we conducted a benchmarking experiment using the meta-evaluation methodology described in~\cite{hedstrom2022quantus} with the corresponding MetaQuantus library~\cite{hedstrom2022quantus}.  

\subsubsection{Methodology.} To measure metric reliability, two steps are followed. First, two types of controlled perturbation are generated: minor and disruptive, respectively. These perturbations capture the metric's resilience to noise ($NR$) and reactivity to adversary ($AR$), respectively. Effectively, these perturbations are applied to the input- and model spaces, separately, creating two tests, i.e., the Input Perturbation Test (IPT) and the Model Perturbation Test (MPT)\footnote{For the IPT, we applied i.i.d additive uniform noise such that $\hat{\vx}_i = \vx + \bm{\delta}_i \ \text{where} \ \bm{\delta}i \sim {\mathcal{U}}(\alpha, \beta)$. For the MPT, we applied multiplicative Gaussian noise to all weights of the network, i.e., $\hat{\bm{\theta}}i = \bm{\theta} \cdot \bm{\nu}i \ \text{where} \ \bm{\nu}i \sim \mathcal{N}(\bm{\mu}, \bm{\Sigma})$. The hyperparameters $\alpha, \beta, \mu, \Sigma$ were set according to the original publication~\cite{hedstrom2023metaquantus}.}. Second, the effects of the perturbations are measured in two meta-evaluative criteria: intra-consistency ($\mathbf{IAC}$) and inter-consistency ($\mathbf{IEC}$). Here, $\mathbf{IAC}$ refers to measuring the similarity in score distributions post-perturbation and $\mathbf{IEC}$ refers to the occurrence of categorical ranking changes within a set of distinct explanation methods\footnote{Here, $\mathbf{IAC}$ returns a normalised p-value from the non-parametric \textit{Wilcoxon signed-rank test}\cite{wilcoxon1945} comparing score distributions of the original and perturbed evaluation case, where similar distributions are expected for the $NR$ case and dissimilar distributions are expected for the $AR$ case. $\mathbf{IEC}$ counts the changes in categorical rankings within a set of explanation methods, post-perturbation, where an optimal-performing metric generates the same ranking for minor perturbation, i.e., the $NR$ case, and lower rankings for disruptive noise, i.e., the $AR$ case. Complete mathematical definitions for contents in $\mathbf{m}$ (Equation \ref{eq:meta-eval-vector}), are found in the original publication\cite{hedstrom2023metaquantus}.}. Each metric consequently receives a summarising meta-consistency score with $ \mathbf{MC}\in [0, 1]$:
\begin{equation}\label{eq:meta-eval-vector}
    \mathbf{MC} = \left(\frac{1}{|\mathbf{m}^{*}|}\right){\mathbf{m}^{*}}^T\mathbf{m} \quad \text{where} \quad \mathbf{m} = \begin{bmatrix} \mathbf{IAC}_{NR}\\ \mathbf{IAC}_{AR}\\ \mathbf{IEC}_{NR}\\     \mathbf{IEC}_{AR} \end{bmatrix},
 \end{equation}
with $\vm^{*} \in \mathbb{R}^4$
represents an optimally performing quality estimator, i.e., an all-one vector. A higher $\mathbf{MC}$ score, approaching 1, demonstrates greater reliability performance based on the tested scoring criteria. 
$\overline{\text{MC}}$ refers to the average $\mathbf{MC}$ score across IPT and MPT.
Further details on experimental hyperparameters are discussed in Supplement \ref{sec:benchmarking_details}.  
We refer to the original publication~\cite{hedstrom2023metaquantus} for further details on the framework.

\begin{figure}[!t]
    \centering
    \includegraphics[width=\linewidth]{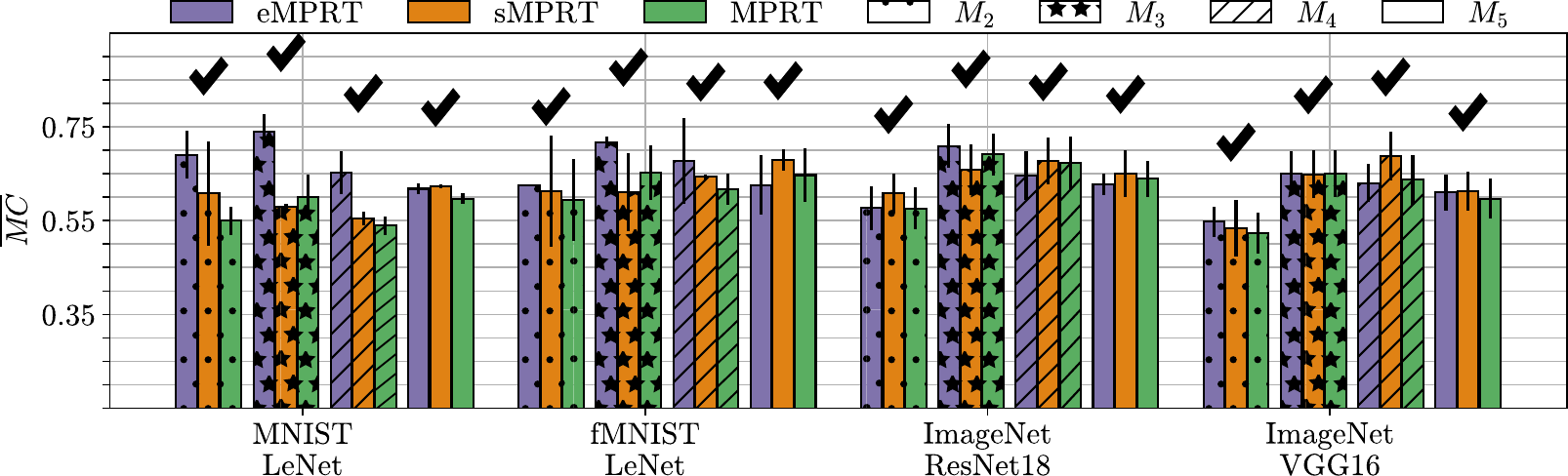}
    \captionsetup{font=footnotesize}
    \caption{Summary of average $\overline{\text{MC}}$ scores for each metric across various tasks, compiled from $3$ iterations with $K=5$ perturbations (detailed in Table \ref{table-benchmarking}). The groups $M_2,$ $M_3,$ $M_4,$ and $M_5$ correspond to distinct sets of XAI methods, where $M_2=${\textit{GradientSHAP, IntegratedGradients}}, $M_3=${\textit{Saliency, LRP-${z^+}$, Input$\times$Gradient}}, $M_4=${\textit{Gradient, GradCAM, LRP-$\varepsilon$, Guided-Backprop}}, and $M_5=${\textit{Guided-Backprop, GradientSHAP, GradCAM, LRP-$\epsilon$, Saliency}}. The colour intensity on the figure represents the specific task, reflecting different dataset and model combinations.}
    \label{fig:average-mc}
    \vspace{-0.75em}
\end{figure}

\begin{table}[t!]
    \centering
    \captionsetup{font=footnotesize}
    \caption{Consolidated results of the meta-evaluation, after $3$ iterations with $K=5$. The metric with the best performance in each scenario is emphasised in bold. If this metric surpasses the second best by a minimum of one standard deviation, it is \underline{underlined}. Higher scores are desirable.}
    \label{table-benchmarking}
\begin{tabular}{ccccc}

\toprule
          \textbf{Setting} &                  \textbf{XAI Methods} &             \textbf{eMPRT} ($\uparrow$) &             \textbf{sMPRT} ($\uparrow$) &              \textbf{MPRT} ($\uparrow$) \\
\midrule
\multirow{4}{*}{\texttt{fMNIST; LeNet}} &              \textit{$M_2$} & \textbf{0.625 $\pm$ 0.002} & 0.613 $\pm$ 0.119 & 0.594 $\pm$ 0.088 \\
\cline{3-5}
 &     \textit{$M_3$} & \textbf{\underline{0.717 $\pm$ 0.012}} &  0.610 $\pm$ 0.083 & 0.653 $\pm$ 0.058 \\
 \cline{3-5}
 &     \textit{$M_4$} & \underline{\textbf{0.677 $\pm$ 0.091}} & 0.643 $\pm$ 0.005 & 0.617 $\pm$ 0.033 \\
 \cline{3-5}
 &     \textit{$M_5$} & 0.626 $\pm$ 0.064 &  \underline{\textbf{0.680 $\pm$ 0.023}} & 0.647 $\pm$ 0.057 \\
 \midrule
\multirow{4}{*}{\texttt{MNIST; LeNet}} &              \textit{$M_2$} &  \underline{\textbf{0.690 $\pm$ 0.051}} & 0.608 $\pm$ 0.111 &   0.550 $\pm$ 0.030 \\
\cline{3-5}
 &     \textit{$M_3$} & \underline{\textbf{0.740 $\pm$ 0.036}} &  0.580 $\pm$ 0.006 &   0.600 $\pm$ 0.048 \\
 &     \textit{$M_4$} & \underline{\textbf{0.653 $\pm$ 0.046}} & 0.555 $\pm$ 0.015 & 0.539 $\pm$ 0.019 \\
 \cline{3-5}
 &     \textit{$M_5$} & 0.618 $\pm$ 0.012 & \textbf{0.623 $\pm$ 0.004} & 0.597 $\pm$ 0.012 \\
  \midrule
\multirow{4}{*}{\texttt{ImageNet; ResNet18}} &              \textit{$M_2$} & 0.577 $\pm$ 0.047 & \textbf{0.608 $\pm$ 0.042} & 0.576 $\pm$ 0.045 \\
\cline{3-5}
 &     \textit{$M_3$} &   \textbf{0.709  $\pm$ 0.047} &  0.658 $\pm$ 0.054 & {{0.691 $\pm$ 0.045}} \\
 \cline{3-5}
 &     \textit{$M_4$} & 0.646 $\pm$ 0.052 &  \textbf{0.678 $\pm$ 0.050} & 0.673 $\pm$ 0.057 \\
 \cline{3-5}
 &     \textit{$M_5$} & 0.628 $\pm$ 0.023 &  \textbf{0.651 $\pm$ 0.050} & 0.639 $\pm$ 0.039 \\
 \midrule
\multirow{4}{*}{\texttt{ImageNet; VGG16}} &              \textit{$M_2$} & \textbf{0.548 $\pm$ 0.032} & 0.533 $\pm$ 0.06 & 0.523 $\pm$ 0.043 \\
\cline{3-5}
 &     \textit{$M_3$} & \textbf{0.651 $\pm$ 0.047} & 0.649 $\pm$ 0.051 &   0.650 $\pm$ 0.050 \\
 \cline{3-5}
 &     \textit{$M_4$} &  0.630 $\pm$ 0.041 & {\textbf{0.688 $\pm$ 0.052}} & 0.637 $\pm$ 0.053 \\
 \cline{3-5}
 &     \textit{$M_5$} &  \textbf{0.610 $\pm$ 0.039} & 0.613 $\pm$ 0.041 & 0.597 $\pm$ 0.043 \\
\bottomrule
\end{tabular}
\end{table}

\subsubsection{Results.} \label{sec:emprt-results} As Figure \ref{fig:average-mc} displays, MPRT is generally outperformed by any of its variants (eMPRT or sMPRT) across the tested tasks and attribution sets $\{M_2, \cdots, M_5\}$ (indicated symbols over grouped bars). These results are encouraging, as they suggest that the modifications made to MPRT have rendered the original metric more reliable. However, neither Figure \ref{fig:average-mc} nor Table \ref{table-benchmarking} indicates perfect reliability, i.e., $\overline{\text{MC}} = 1$, for any metric variant. This highlights the ongoing necessity for careful use of evaluation metrics, particularly in the field of XAI where definitive ground truth explanation labels are often unavailable. While eMPRT and sMPRT show superior performance, not all experimental settings yield statistically significant differences, as detailed in Table \ref{table-benchmarking}\footnote{Overlapping scores between metric variants are expected given the similarity of the metrics (Definitions \ref{def-mprt}-\ref{def-emprt}), and do not indicate that further repetitions ($K$ and $iters$) are necessary. The average MC score standard deviations across this set are comparatively small and do not change significantly across XAI groups, which indicates results stability.}. Therefore, a nuanced, holistic approach to adopting metrics in XAI is critical. Further benchmarking results are provided in Supplement \ref{sec:benchmarking-extra-results}.

\section{Discussion}

In this work, we introduced two extensions to the MPRT: sMPRT and eMPRT. These extensions address key limitations in the original MPRT definition, i.e., the usage of pairwise similarity measures and top-down order of layer randomisation. Our experiments demonstrated enhanced reliability across a wide range of datasets and models. In the following, we discuss the limitations of proposed approaches, outline recommendations for explanation methods and describe future work.

\subsection{Limitations}

Despite sMPRT's advantage of being more algorithimcally similar to the original MPRT and thus can be functionally interpreted in the same way, sMPRT has several limitations. First, it is more computationally expensive than standard MPRT, as additional sampling is required ($N \geq 50$, see Figure \ref{fig:smprt_n}). Second, a defining characteristic of some attribution methods such as \textit{SmoothGrad}~\cite{smilkov2017smoothgrad} and \textit{NoiseGrad}~\cite{bykov2021noisegrad} is their ability to reduce attribution noise by adding noise, limiting sMPRT's efficacy with these methods and blurring the distinction with their baseline methods. Third, the degree of noisiness is an arguable property of attribution methods and removing it before evaluation may yield non-representative or biased results which may create performance discrepancies upon real-world application. Lastly, sMPRT introduces additional hyperparameters $\sigma$ and $N$. These parameters can be chosen depending on context and may not be tunable on any given data domain, e.g., climate data~\cite{bommer2023}, although we found that generally, a larger $N$ yields a better denoised estimate (see Figure \ref{fig:smprt_n}).

eMPRT is generally is more computationally efficient compared to MPRT, since explanations are only computed twice, i.e., pre- and post-model randomisation. It further avoids issues related to layer-order as well as the choice of normalisation function (since the chosen entropy function already normalises implicitly, see Equation \ref{eq:hist_entropy}).
Most importantly, by incorporating the initial unperturbed complexity estimate, i.e., $\xi(\ve)$ into Equation \ref{eq:q_emprt}, eMPRT anchors its quality estimate by the explanation method's inherent complexity, which makes the evaluation scores more comparable across explanation methods and models. eMPRT may, however, exhibit variability based on different tasks or data-related factors, such as object size in vision datasets. 
The complexity metric also introduces the number of bins $B$ (Equation \ref{eq:hist_entropy}) as an additional hyperparameter, which may need adaptation depending on the task or dimensionality of explanations.

\subsection{Recommendations on Attribution Methods}

The original work \cite{adebayo2018} recommends gradient-based methods over modified backpropagation approaches, as the former methods were found to be more sensitive to model parameter changes. However, our findings suggest a different perspective on these claims where, e.g., removing noise from explanations impacts various attribution methods differently, with gradient-based methods experiencing the most significant degradation (see Figure \ref{fig:smprt}). Additionally, the order of layer randomisation influences how MPRT ranks attribution methods (elaborated in Supplement \ref{sec:layerorder_supplement}) and comparisons to eMPRT rankings yield divergent outcomes (Section \ref{sec:compare-rankings}).

All evaluated methods scored poorly in absolute term for sMPRT and eMPRT, with small margins between XAI methods. However, all tested methods exhibited \emph{some degree} of change with model randomisation, indicating that they all pass the binary test of sensitivity to model parameters. That said, no evaluation metric in isolation is sufficient to determine explanation quality. Instead, as stated in Section \ref{sec:emprt-results}, it is critical to conduct a comprehensive assessment of multiple evaluation criteria and metrics, yielding a more holistic view of the strengths and weaknesses of the individual methods.

\subsection{Future work}

Considering the distinct underlying mechanisms of sMPRT and eMPRT, there is potential for a more comprehensive evaluation approach by combining them, an avenue we intend to investigate in future research. Furthermore, while we present an initial examination into the impacts of layer-order randomisation (see Supplement \ref{sec:layerorder_supplement}), further exploration is needed. Specifically, additional empirical studies are necessary to validate the application of bottom-up randomisation.

\begin{credits}
\subsubsection{\ackname}
Supported by BIFOLD (ref. 01IS18025A and ref. 01IS18037A); 
the European Union’s Horizon Europe research and innovation programme (EU Horizon Europe) as grant TEMA (101093003); 
the German Federal Ministry for Education and Research through project Explaining 4.0 (ref. 01IS200551);
the Investitionsbank Berlin through BerDiBa (grant no. 10174498); 
and the European Union’s Horizon 2020 research and innovation programme (EU Horizon 2020) as grant iToBoS (965221).
\subsubsection{\discintname}
The authors have no competing interests to declare that are
relevant to the content of this article.
\end{credits}

\bibliographystyle{splncs04}
\bibliography{main_20240503174405}

\newpage

\setcounter{figure}{0}
\renewcommand{\thefigure}{S.\arabic{figure}}

\setcounter{section}{0}
\setcounter{subsection}{0}
\renewcommand{\thesubsection}{S.\arabic{subsection}}

\section*{Supplementary Material}

Supplementary Material first provides a detailed explanation of the experimental hyperparameters used in the main paper. It also explores several additional experimental results, offering supporting evidence.

\subsection{Details on Experiments}
\label{sec:experimental_setup_supplement}

\subsubsection{Software Environment.} We implemented all experiments in python, using PyTorch~\cite{Pytorch2019} for deep learning. To compute explanations, we utilised the zennit~\cite{Anders2021Software} and captum~\cite{Kokhlikyan2020Captum} libraries. The Quantus~\cite{hedstrom2022quantus} package was used to evaluate with MPRT. We further extended this package to implement sMPRT and eMPRT. The MetaQuantus library was employed for meta-evaluation~\cite{hedstrom2023metaquantus}.

\subsubsection{Data.}
Existing research~\cite{yona2021,kokhlikyan2021investigating} shows that MPRT results can vary depending on the task. To yield more robust results, we therefore include a range of datasets in this work.
Specifically, we make use of the following three image classification datasets: ImageNet (ILSVRC2012~\cite{Krizhevsky2012Imagenet}), MNIST\cite{lecun2010mnist} and fMNIST~\cite{fashionmnist2015}. We randomly sample $1000$ test samples for fMNIST and MNIST. For ImageNet we utilise the $1000$ first samples of the validation set for experiments discussed in Figure \ref{fig:smprt} and randomly select $300$ test samples for experiments showcased in Figure \ref{fig:emprt} and Figures \ref{fig:average-mc}-\ref{fig:metaquantus-$M_5$}.

\subsubsection{Models.}
The experiments are conducted with a variety of neural network architectures, including LeNets~\cite{lecun1998gradient}, ResNets~\cite{he2015deep}, and VGGs~\cite{simonyan2014very}, to ensure consistency of our findings. We train LeNet models to achieve accuracies of $98.14\%$ and $87.44\%$, respectively for MNIST and fMNIST. The training process for all models follows a similar procedure: SGD optimization with an initial learning rate of $0.001$, a momentum of $0.9$, and a standard cross-entropy loss. $20$ epochs is used for training. For ILSVRC2012~\cite{Krizhevsky2012Imagenet}%
, we use the ResNet-18~\cite{he2015deep} and VGG-16~\cite{simonyan2014very} models with pre-trained ImageNet weights. These models and weights were accessed through the PyTorch library \cite{Pytorch2019}.

\subsubsection{Explanation Methods and Preprocessing.} \label{preprocess}
Our evaluation included several explanation methods:
\textit{Gradients}~\cite{morch,baehrens}, \textit{Saliency}~\cite{Simonyandeep}, \textit{Input$\times$Gradient}~\cite{ShrikumarInputXG}, GradCAM~\cite{selvaraju2019}, \textit{GradientSHAP}~\cite{lundberg2017unified} with $5$ samples, \textit{SmoothGrad}~\cite{smilkov2017smoothgrad} (using 20 noisy samples and a noise level of $0.1 \slash (x_\text{max}-x_\text{min})$; these are the default parameters in zennit~\cite{Anders2021Software}, \textit{Integrated Gradients}~\cite{sundararajan2017axiomatic} (with 20 iterations and a baseline of zero), \textit{Guided Backpropagation}~\cite{springenberg2014striving} and two different variations of \textit{Layer-wise Relevance Propagation} (LRP)~\cite{bach2015pixel,montavon2017dtd}: application of the \textit{LRP-$\varepsilon$-rule}~\cite{bach2015pixel} with $\varepsilon = 1e^{-6}$ to all layers (called LRP-$\varepsilon$) and application of the \textit{LRP-${z^+}$-rule}~\cite{montavon2017dtd} to all layers (called LRP-$z^+$). We also utilised random attributions (as a control variant); these were sampled from a uniform distribution, i.e., $\hat{\ve}_i \sim \mathcal{U}(1, 0)$.
Since preprocessing can influence the results of the MPRT~\cite{sundararajan2018}, we apply specific preprocessing steps based on the explanation method. For methods where the sign of the attribution does not carry meaning in terms of feature importance, such as \textit{Saliency} and \textit{SmoothGrad}, we take the absolute values of the attributions. For \textit{GradCAM}, only positive values were considered, as outlined in the original publication \cite{selvaraju2019}. For all other methods, we do not alter the signs of values. Although the complexity measure used in eMPRT inherently normalises inputs, the SSIM, used to assess explanation differences in MPRT and sMPRT, requires prior normalisation to ensure comparability of attributions at different scales before and after randomisation. To achieve this, we applied normalisation using the square root of the average second-moment estimate (as elaborated in Supplements of~\cite{bindershort}). This approach introduces less additional variance compared to e.g., normalisation by the maximum value. The normalisation method is defined as follows:
\begin{equation}
\label{eq:normalise-eq}
   \text{norm}(\ve) = \frac{{\ve}_{h,w}}{\left(\frac{1}{HW}\sum_{h',w'}{\ve}_{h',w'}^2\right)^{1/2} }~,
\end{equation}
where $H$, $W$ is the height and width, respectively, and $\hat{\ve}_{h,w}$ denotes the explanation value at pixel location ($h$, $w$) 
\footnote{This normalisation makes sure that the mean squared distance of each explanation score from zero is one. Unlike other normalisations, this technique does not restrict the attribution values to a fixed range, which makes it unsuitable for visualization purposes, where a fixed range is generally preferred; instead it preserves a measure useful for comparing distances between different explanation methods.}. 

\subsubsection{MPRT Hyperparameters.}
The original paper~\cite{adebayo2018}, uses different similarity measures such as SSIM~\cite{nilsson2020understanding}, \textit{Spearman Rank Correlation} and \textit{HOG}. We employ \textit{SSIM}~\cite{nilsson2020understanding} due to its widespread adoption in existing literature~\cite{adebayo2018,sixt2019,bindershort}.

\subsubsection{sMPRT Hyperparameters.}
We employ \textit{SSIM}~\cite{nilsson2020understanding} to the similarity between explanations in sMPRT. For generating noise ($\sigma$), we utilise an adaptive standard deviation of $0.2 \slash (x_\text{max}-x_\text{min})$, following the original heuristic~\cite{smilkov2017smoothgrad}. We explored the following values for the number of noise samples $N$: $1, 20, 50, 300$. As we show in Figure \ref{fig:smprt_n}, larger $N$ generally lead to better estimates of denoised explanations, but also increases computation time. In Section \ref{sec:smprt}, we therefore chose $N=50$ as a relatively accurate estimate that still has acceptable runtime.

\subsubsection{eMPRT Hyperparameters.} \label{emprt_hyper}

We set the bin count (cf. Equation \ref{eq:hist_entropy}) empirically to $B=100$ in our experiments, as it showed robust performance across various tasks and attribution methods. Although there exist various statistical rules such as \textit{Freedman-Diaconis'}~\cite{freedman1981histogram} and \textit{Scotts'}~\cite{scott1979optimal} for the optimisation of $B$, initial experiments demonstrate inconsistencies in results. These rules make assumptions such as data normality, which does not apply universally across XAI methods.

To measure the complexity of model outputs, we computed the Shannon entropy of the post-softmax probabilities as follows:
\begin{equation} \label{shannon-entropy}
\xi(f_L(\vx)) = -\sum_{i=1}^{|L|} p(x_i) \log_2(p(x_i))
\end{equation}

\subsubsection{Meta-Evaluation Details.}
\label{sec:benchmarking_details}

For empirical assessment, we utilised the pre-existing tests available in the MetaQuantus library~\cite{hedstrom2022quantus} and their associated hyperparameters. The existing MetaQuantus tests are accessible via the GitHub repository\footnote{Code at: \url{https://github.com/annahedstroem/MetaQuantus/} and hyperparameter settings at notebook: \url{https://github.com/annahedstroem/MetaQuantus/blob/main/tutorials/Tutorial-Getting-Started-with-MetaQuantus.ipynb}.}. All metrics have been implemented in Quantus~\cite{hedstrom2022quantus}. We executed these metrics over $K=5$ perturbations, spanning $3$ iterations with the test configurations as outlined in the notebook for four different sets of explanation methods. Here, XAI groups were formed by randomly selecting XAI methods from the full set of methods, maintaining consistency across experimental settings such as dataset and model combinations. Moreover, Appendix A.5.2 in the original publication~\cite{hedstrom2023metaquantus} demonstrates that the sets of explanation methods exert limited impact on the average MC score. Regarding the selection of $K$ and the number of iterations, we adhered to guidance from the original publication~\cite{hedstrom2023metaquantus} to ensure that the standard deviation between the sets is comparatively low. Furthermore, to ensure a fair comparison among the metrics, we adhered to uniform hyperparameter settings, as specified by the normalisation formula in Equation \ref{eq:normalise-eq}. Given that the MetaQuantus library necessitates each metric to yield a single quality estimate $\hat{q} \in \mathbb{R}$ per sample, we calculated the correlation coefficient for the fully randomised models in both MPRT and sMPRT, which is anticipated to enhance their performance. 

\subsection{Sensitivity of MPRT to Layer Randomisation Order}
\label{sec:layerorder_supplement}

\begin{figure}[!b]
    \centering
    \includegraphics[width=0.95\linewidth]{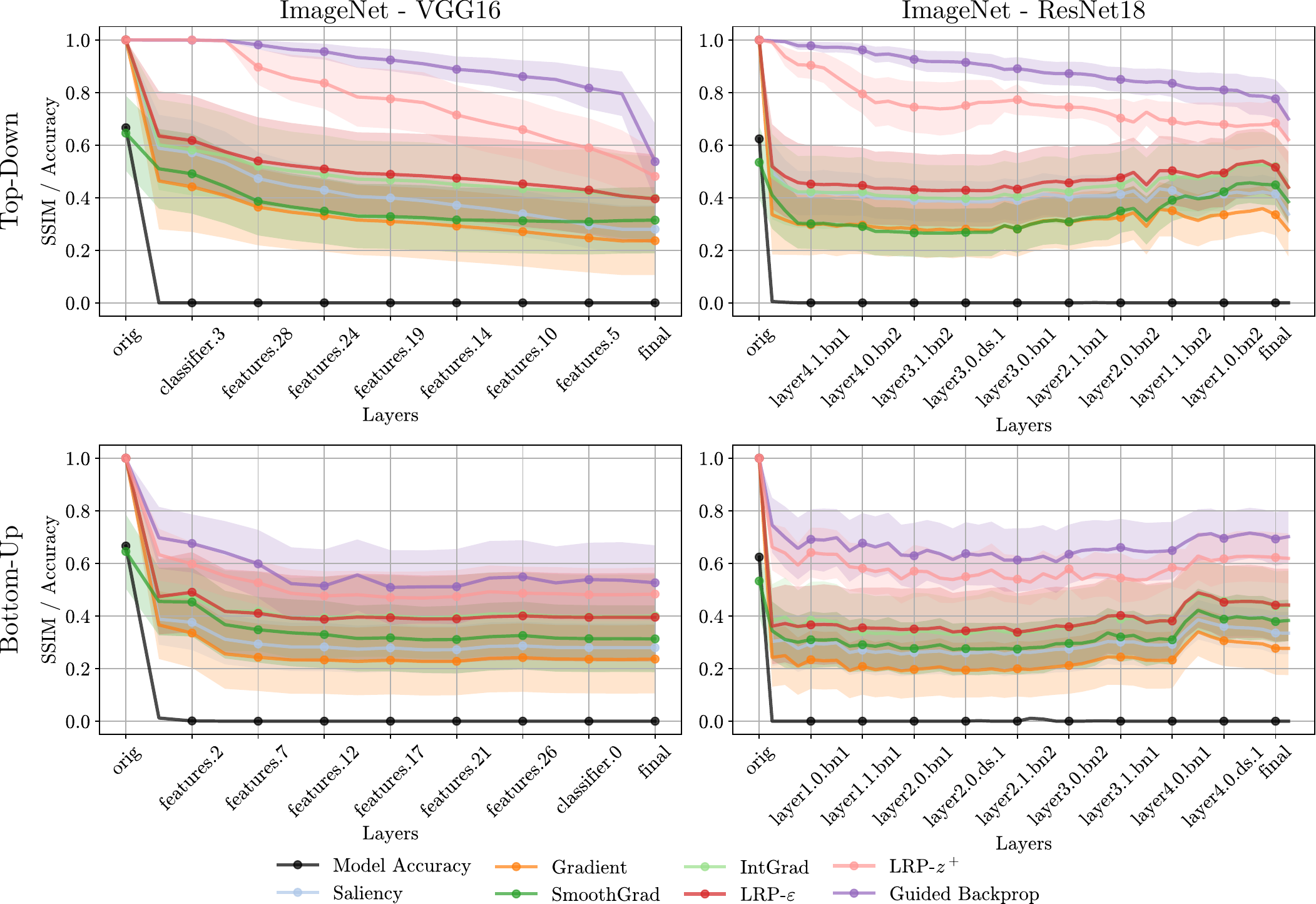}
    \captionsetup{font=footnotesize}
    \caption{Layer randomisation order affects MPRT results. Results are plotted for ImageNet, as well as VGG16 \emph{(left)} and ResNet18 \emph{(right}). \emph{(Black line)} change in model accuracy with randomisation; \emph{(Other lines)} \textit{SSIM} score from MPRT for several explanation methods. The unrandomised and fully randomised model states are denoted as \emph{orig} and \emph{final}, respectively. \emph{(Top)} Randomisation in top-down order, i.e., from highest to lowest layer, as employed by the authors of~\cite{adebayo2018}. \emph{(Bottom)} Randomisation in top-down order, from lowest to highest layer, as suggested by recent work~\cite{bindershort}. For both order types, model accuracy decreases with the same speed. Yet, we observe a large difference in explanation \textit{SSIM} scores between layer-orders: for top-down order, there is a higher discrepancy in faithfulness between explanation methods, as measured by the MPRT. }
    \label{fig:mprt_layerorder}
    \vspace{-0.5em}
\end{figure}

To prevent the preservation of significant portions of the forward pass, which can happen with top-down randomisation (as discussed in Section \ref{issues} (b) and in recent work~\cite{bindershort}), the authors of~\cite{bindershort} suggest randomising layers in bottom-up order. This approach avoids such preservation because it disrupts the lower layers before the higher ones.
Given that we use bottom-up randomisation in our experiments on sMPRT and that this approach has yet to be empirically validated, we investigate how it affects MPRT results in the following.
However, it's worth noting that a potentially more straightforward alternative is to avoid layer-wise randomisation entirely and instead focus on comparing the original (unrandomised) model with a fully randomised model. This approach, outlined in Equation \ref{eq:q_emprt}, might offer a more consistent evaluation by sidestepping unwanted potential side-effects of partial randomisation.

Figure \ref{fig:mprt_layerorder} illustrates a comparison between MPRT rankings using two different randomisation orders: top-down \emph{(top)} and bottom-up \emph{(bottom)}. Our approach differs from the the seminal work~\cite{adebayo2018} in the applied preprocessing step for the explanations, where we normalise using the recommended approach in recent work~\cite{bindershort} (also cf. discussion in Section \ref{issues} (a)), i.e., the square root of the average second-moment estimate. We only consider the absolute values for \textit{Saliency} and \textit{SmoothGrad}.

We first confirm that the model is sufficiently disrupted by both top-down and bottom-up randomisation by tracking the model's accuracy. In all four settings, the model accuracy (\emph{black} lines) drops to random levels for the ImageNet experiments (approximately $0.001$) immediately after randomising the first layer.
However, the results of the MPRT vary considerably with the order of randomisation. Specifically, all explanation methods perform better with bottom-up randomisation. Additionally, the SSIM tends to drop more rapidly with bottom-up randomisation, leading to smaller differences in SSIM scores between explanation methods. Notably, explanation methods that seemed almost unaffected by model parameters with top-down randomisation---such as \textit{LRP-$z^+$} and \textit{Guided Backpropagation}---show greater faithfulness to the model's parameters as measured by the MPRT with bottom-up randomisation.

\begin{figure}[!t]
    \centering
    \includegraphics[width=0.95\linewidth]{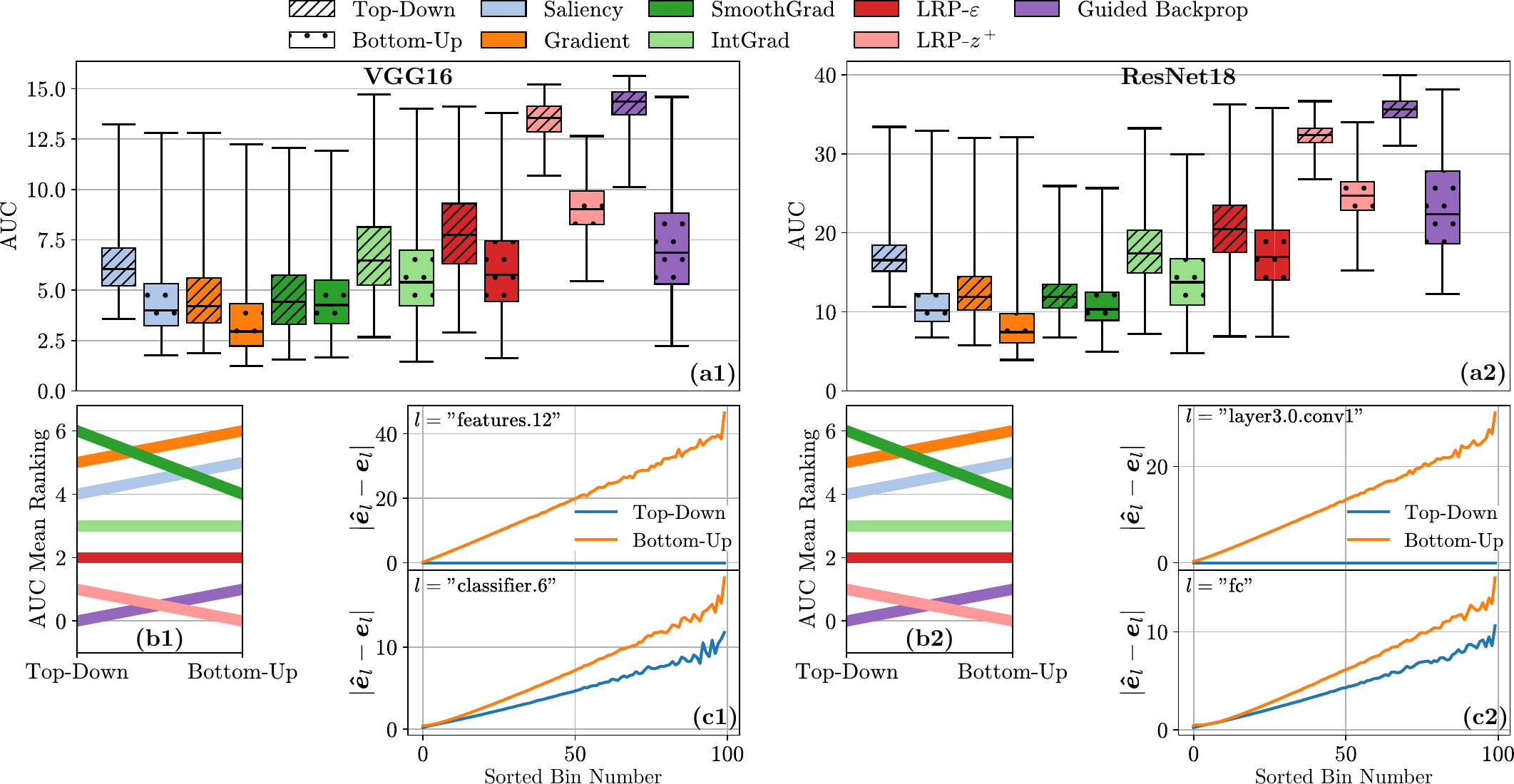}
    \captionsetup{font=footnotesize}
    \caption{Layer randomisation order affects MPRT results. We investigated VGG16 \emph{(left)} and ResNet18 \emph{(right)} models on the ImageNet dataset. \emph{a1 and a2}: Area under MPRT curves (cf. Figure \ref{fig:mprt_layerorder}) for various XAI methods using top-down and bottom-up randomisation orders, visualised as percentile (0, 25, 50, 75, 100) statistics. \emph{b1 and b2}: Relative ranking changes between attribution methods when switching from top-down to bottom-up order. A lower score in (a) corresponds to a higher ranking in (b). \emph{(c1 and c2)}: Intermediate explanation values $\ve$ (cf.~\cite{bindershort}), sorted into $100$ bins according to their absolute value (x-axis) vs. the average amount of change with randomisation for values in each bin (y-axis).} 
    \label{fig:topdown_vs_bottomup}
\end{figure}

The previous observations are visually supported by Figure \ref{fig:topdown_vs_bottomup} \emph{(a1) and (a2)}, which illustrates that explanation methods generally perform better under the MPRT with bottom-up randomisation (evidenced by the lower dotted hatch bars compared to the striped bars). However, this observation does not necessarily translate into clear categorical rankings between explanation methods \emph{(b1) and (b2)}, which seem only slightly affected by the randomisation order, potentially due to additional factors influencing the results, such as the sensitivity of SSIM to noise (cf. Section \ref{issues}(c), and~\cite{bindershort}). These factors likely introduce confounding effects in the bottom-up randomisation results and are separately addressed by the sMPRT and eMPRT metrics.

To understand why bottom-up randomisation generally results in better MPRT performance, we examine the intermediate (LRP-$\varepsilon$) explanations $\ve_l$. $\ve_l$ explains the $l$-th layer output, in contrast to the model output (corresponding to the last layer, $L$). The original explanation values are sorted into $100$ equidistant bins based on their magnitude. By then applying bottom-up or top-down randomisation (on the first or last parameterised layer only, respectively), we obtain randomised explanations $\hat{\ve}_l$.

For each bin, we calculate the average change in the explanation values under both bottom-up and top-down randomisation (i.e., the average of $|\hat{\ve}_l-\ve_l|$ for each bin). Figure \ref{fig:topdown_vs_bottomup} \emph{(c1) and (c2)} present the results for two different layers $l$---one intermediate layer \emph{(top)} and the output layer \emph{(bottom)} for each model. This analysis reveals that bottom-up randomisation (orange line) disrupts both the intermediate and final layer explanations, whereas top-down randomisation (blue line) only affects the last layer explanation, and with less effect compared to bottom-up randomisation. This suggests that bottom-up randomisation has a more pronounced effect on the model's reasoning. A robust MPRT should fully randomise the model output as we can only then expect faithful explanations to change~\cite{bindershort}. The observed results suggest that bottom-up randomisation is more effective in achieving this.

\subsection{sMPRT: Raw Plots}
\label{sec:smprt-raw}

\begin{figure}[h]
\centering    \includegraphics[width=\linewidth]{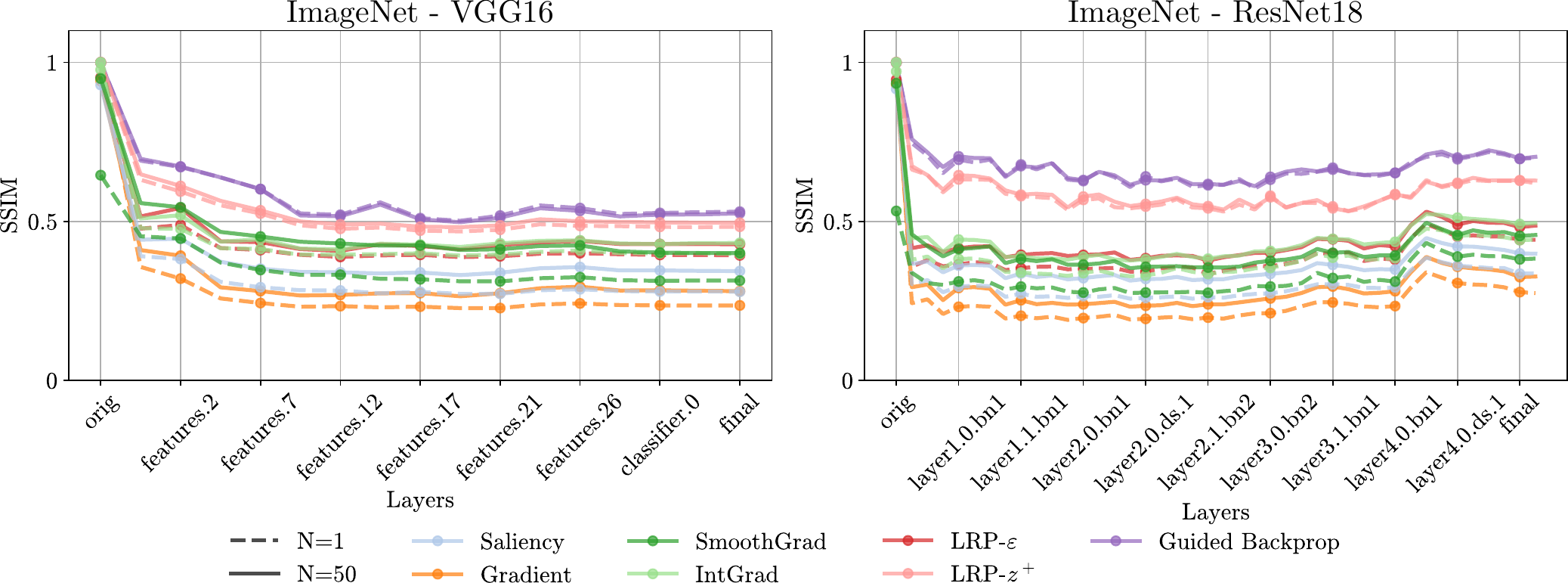}
    \captionsetup{font=footnotesize}
    \caption{sMPRT ($N=50$) versus MPRT ($N=1$) results. Complements Figure \ref{fig:smprt}}
    \label{fig:smprt-data}
    \vspace{-0.5em}
\end{figure}

Figure \ref{fig:smprt-data} complements Figure \ref{fig:smprt}, displaying the raw data.

\subsection{Benchmarking: Additional Details} \label{sec:benchmarking-extra-results}

\begin{figure}[!h]
    \centering
    \includegraphics[width=0.8\linewidth]{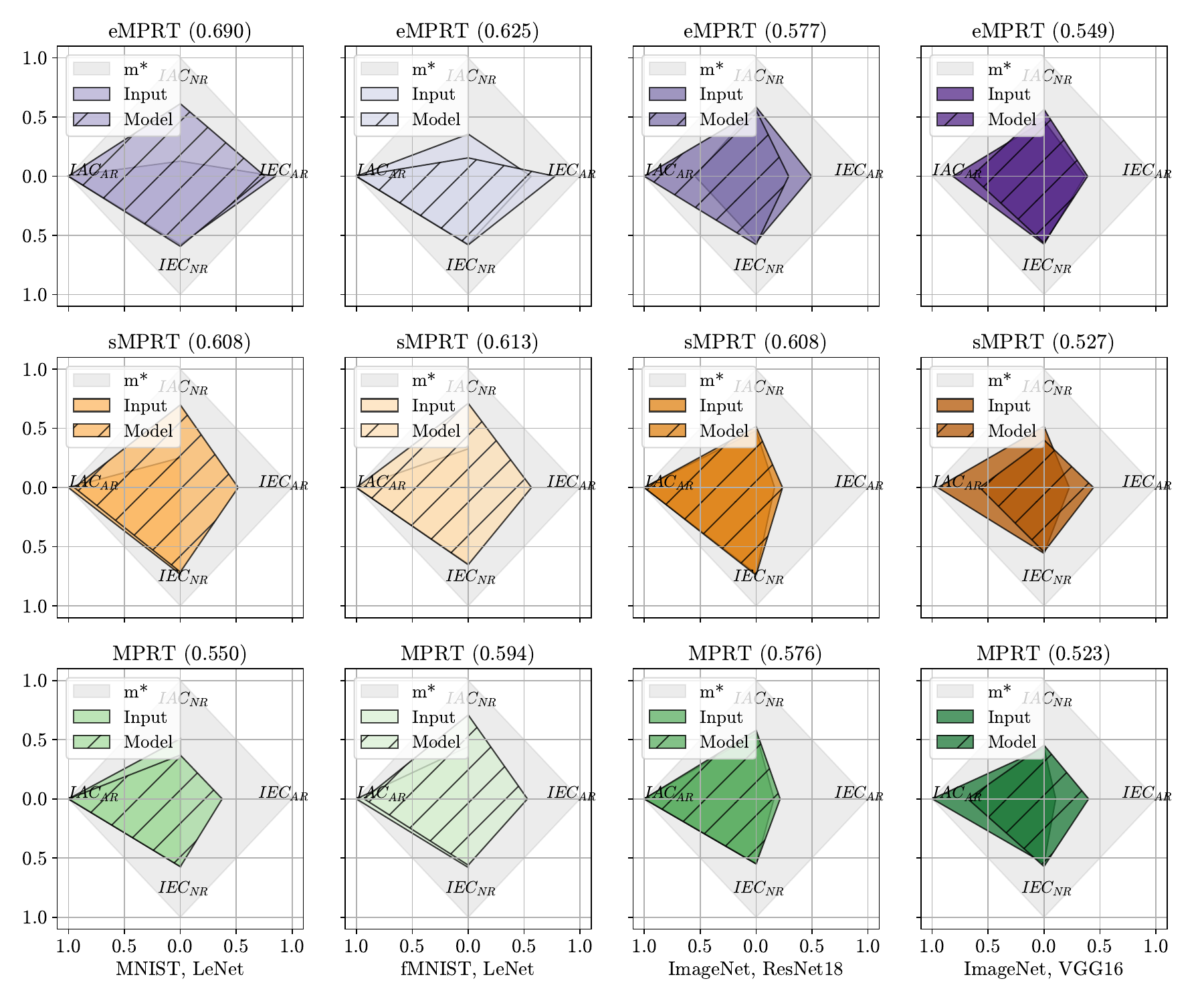}
    \captionsetup{font=footnotesize}
    \caption{Benchmarking outcomes for $M=2$ for MPRT, sMPRT, and eMPRT, with $K=5$ and $3$ iterations. The MC score (in brackets) is averaged over IPT and MPT, with higher scores being better. The optimal metric $\mathbf{m}^{*} = \mathbb{R}^4$ is displayed in the gray.}
    \label{fig:metaquantus-$M_2$}
     \vspace{-0.5em}
\end{figure}

\begin{figure}[!t]
    \centering
    \includegraphics[width=0.8\linewidth]{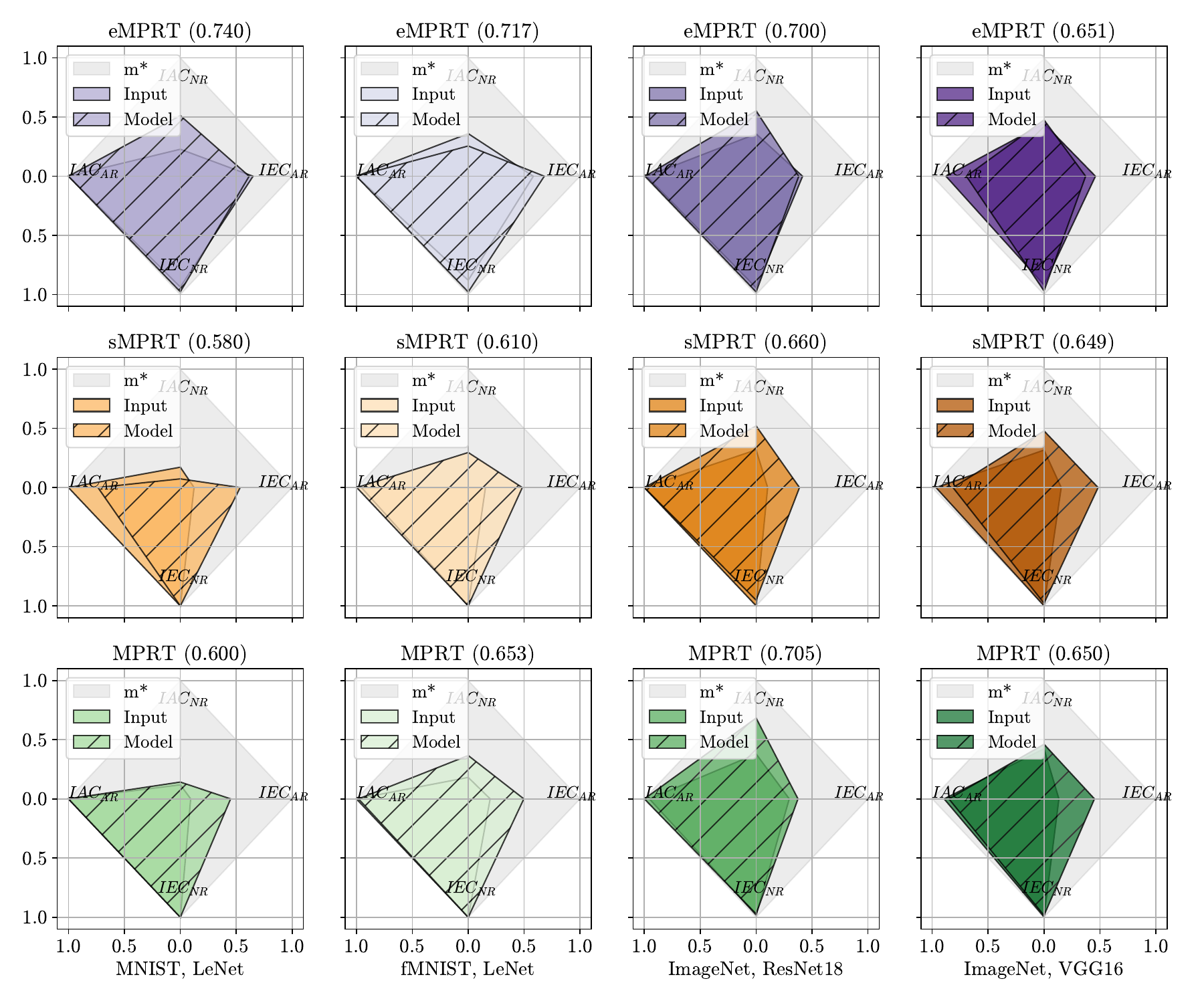}
    \captionsetup{font=footnotesize}
    \caption{Benchmarking outcomes for $M=3$ for MPRT, sMPRT, and eMPRT, with $K=5$ and $3$ iterations. The MC score (in brackets) is averaged over IPT and MPT, with higher scores being better. The optimal metric $\mathbf{m}^{*} = \mathbb{R}^4$ is displayed in the gray.}
    \label{fig:metaquantus-$M_3$}
     \vspace{-0.5em}
\end{figure}

\begin{figure}[!t]
    \centering
    \includegraphics[width=0.8\linewidth]{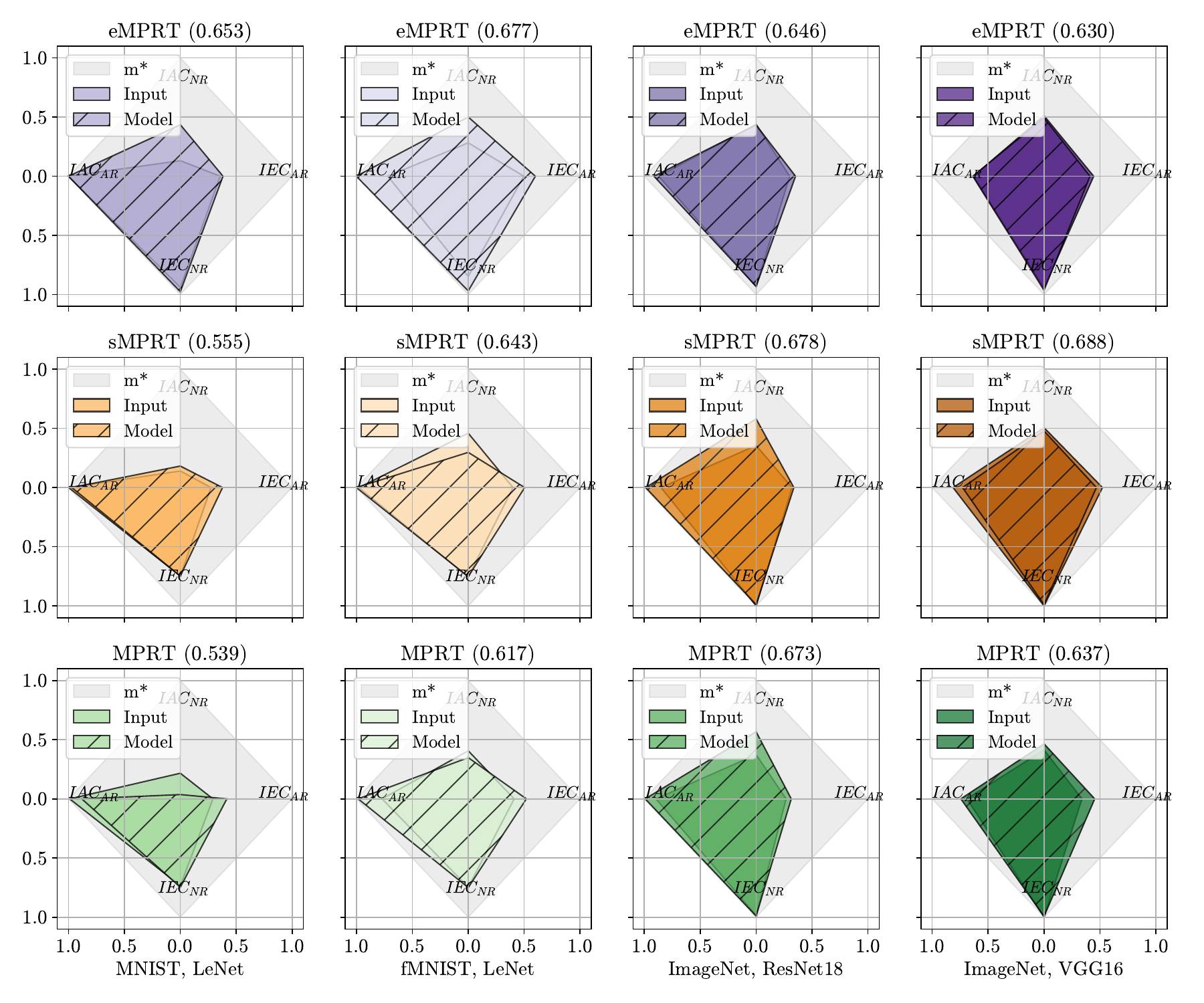}
    \captionsetup{font=footnotesize}
   \caption{Benchmarking outcomes for $M=4$ for MPRT, sMPRT, and eMPRT, with $K=5$ and $3$ iterations. The MC score (in brackets) is averaged over IPT and MPT, with higher scores being better. The optimal metric $\mathbf{m}^{*} = \mathbb{R}^4$ is displayed in the gray.}
    \label{fig:metaquantus-$M_4$}
     \vspace{-0.5em}
\end{figure}

\begin{figure}[!t]
    \centering
    \includegraphics[width=0.8\linewidth]{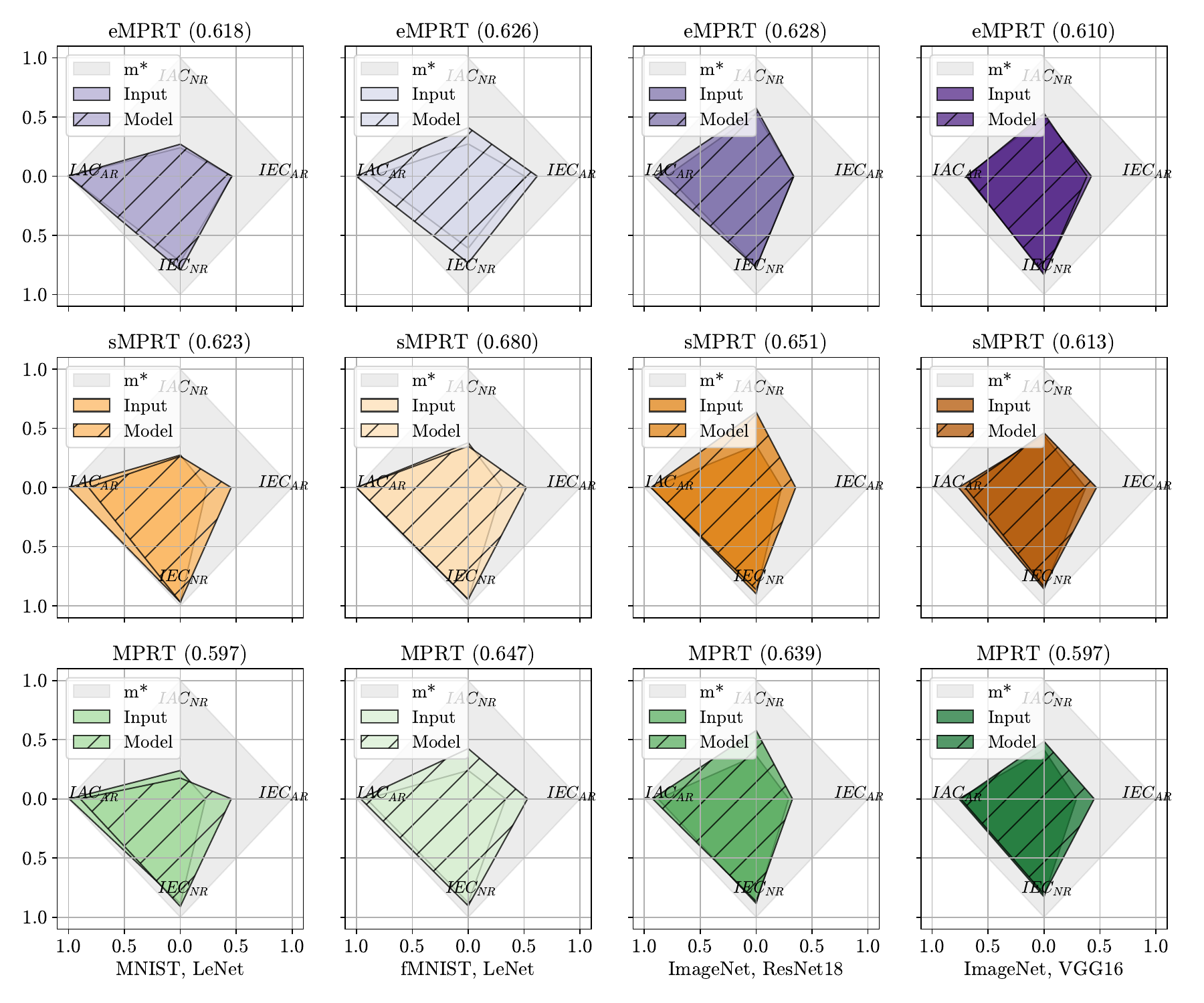}
    \captionsetup{font=footnotesize}
    \caption{Benchmarking outcomes for $M=5$ for MPRT, sMPRT, and eMPRT, with $K=5$ and $3$ iterations. The MC score (in brackets) is averaged over IPT and MPT, with higher scores being better. The optimal metric $\mathbf{m}^{*} = \mathbb{R}^4$ is displayed in the gray.}
    \label{fig:metaquantus-$M_5$}
     \vspace{-0.5em}
\end{figure}

Here, we offer additional analysis of benchmarking outcomes. Figures \ref{fig:metaquantus-$M_2$}-\ref{fig:metaquantus-$M_5$} depict distinct area graphs, mirroring the contents of the meta-evaluation vector. By analysing the size and shape of the coloured regions for each metric variant, we can deduce the performance across tested failure modes ($NR$, $AR$). Larger coloured regions suggest higher metric reliability and the grey area denotes the region of optimal performance for a quality estimator, i.e., $\mathbf{m}^{*} = \mathbb{R}^4$. Consistent colour shades (purple, orange, and green) represent different datasets and are maintained throughout Figures \ref{fig:metaquantus-$M_2$}-\ref{fig:metaquantus-$M_5$} (as well as Figure \ref{fig:average-mc} in the main manuscript). Task descriptions are provided in columns, and metric variations (MPRT, sMPRT, and eMPRT) are arranged in rows. Generally, eMPRT and sMPRT exhibit superior performance across diverse tasks and subsets of attribution methods in both input- and model-perturbation tests, with few exceptions.

\end{document}

%% file: math_commands.tex
\usepackage{amsmath}
\usepackage{amssymb}
\usepackage{amsfonts}
\usepackage{bm}

\def\eqref#1{equation~\ref{#1}}
\def\1{\bm{1}}

\def\ve{{\bm{e}}}

\def\vm{{\bm{m}}}

\def\vx{{\bm{x}}}
\def\vy{{\bm{y}}}

\DeclareMathAlphabet{\mathsfit}{\encodingdefault}{\sfdefault}{m}{sl}
\SetMathAlphabet{\mathsfit}{bold}{\encodingdefault}{\sfdefault}{bx}{n}

\def\sO{{\mathbb{O}}}